\documentclass{article}
\pdfoutput=1 
\usepackage{setspace}
\usepackage{PRIMEarxiv}

\usepackage{amssymb,graphicx,color}
\usepackage{amsfonts}
\usepackage{latexsym}
\usepackage{amsmath}

\usepackage{multirow}
\usepackage[table,xcdraw]{xcolor}
\usepackage{rotating}
\usepackage{tikz}
\usepackage{hyperref}
\usepackage{cleveref}
\usepackage[utf8]{inputenc} 
\usepackage[T1]{fontenc}    
\usepackage{hyperref}       
\usepackage{url}            
\usepackage{booktabs}       
\usepackage{amsfonts}       
\usepackage{nicefrac}       
\usepackage{microtype}      
\usepackage{fancyhdr}       
\usepackage{graphicx}       
\graphicspath{{media/}}     
\usepackage{listings}
\usepackage{xcolor}

\definecolor{codegreen}{rgb}{0,0.6,0}
\definecolor{codegray}{rgb}{0.5,0.5,0.5}
\definecolor{codepurple}{rgb}{0.58,0,0.82}
\definecolor{backcolour}{rgb}{0.95,0.95,0.92}

\lstdefinestyle{mystyle}{
    backgroundcolor=\color{backcolour},   
    commentstyle=\color{codegreen},
    keywordstyle=\color{magenta},
    numberstyle=\tiny\color{codegray},
    stringstyle=\color{codepurple},
    basicstyle=\ttfamily\scriptsize,
    breakatwhitespace=false,         
    breaklines=true,                 
    captionpos=b,                    
    keepspaces=true,                 
    numbers=left,                    
    numbersep=5pt,                  
    showspaces=false,                
    showstringspaces=false,
    showtabs=false,                  
    tabsize=2
}

\title{Automated tabulation of clinical trial results: \\ A joint entity and relation extraction approach with transformer-based language representations}

\author{
  Jetsun Whitton and Anthony Hunter \\
  Department of Computer Science \\
  University College London \\
  Gower Street, London WC1E 6BT, UK\\
  \texttt{\{jetsun.whitton.20, anthony.hunter\}}\\
  \texttt{@ucl.ac.uk}
  }

\begin{document}

\maketitle
\begin{abstract}
Evidence-based medicine, the practice in which healthcare professionals refer to the best available evidence when making decisions, forms the foundation of modern healthcare. However, it relies on labour-intensive systematic reviews, where domain specialists must aggregate and extract information from thousands of publications, primarily of randomised controlled trial (RCT) results, into evidence tables.
 This paper investigates automating evidence table generation by decomposing the problem across two language processing tasks: \textit{named entity recognition}, which identifies key entities within text, such as drug names, and \textit{relation extraction}, which maps their relationships for separating them into ordered tuples. We focus on the automatic tabulation of sentences from published RCT abstracts that report the results of the study outcomes. 
Two deep neural net models were developed as part of a joint extraction pipeline, using the principles of transfer learning and transformer-based language representations. To train and test these models, a new gold-standard corpus was developed, comprising almost 600 result sentences from six disease areas.
 This approach demonstrated significant advantages, with our system performing well across multiple natural language processing tasks and disease areas, as well as in generalising to disease domains unseen during training. Furthermore, we show these results were achievable through training our models on as few as 200 example sentences. The final system is a proof of concept that the generation of evidence tables can be semi-automated, representing a step towards fully automating systematic reviews.

\end{abstract}


\section{Introduction} 
\label{ch:introduction}
Over the last three decades, decision making in clinical practice has been driven by the systematic evaluation of healthcare evidence in what is known as evidence-based medicine (EBM). Defined as: \textit{``the conscientious, explicit and judicious use of current best evidence in making decisions about the care of individual patients."}, EBM sets systematic evaluation standards to reduce bias and improve quality in clinical reports \cite{Sackett71,sackett1995need}. Its goal is to combine high quality evidence with clinical experience and patient preference to achieve the best possible outcomes in care.  

The gold standard for evidence in the healthcare domain is the randomised controlled trial (RCT) -- a study where selected participants are randomly allocated into groups to test a specific drug, treatment or other intervention. These groups, also known as trial arms, are allocated either the study intervention (study arm[s]) or a comparator (control arm[s]) that could be another intervention or placebo. Both arms are then measured and compared over a period of time on a set of predefined outcomes, primarily efficacy and safety, to ascertain the effectiveness of the study intervention. 

In the same time-frame that EBM has become the mainstay of modern medicine, the number of registered clinical trial studies has risen exponentially \cite{TrendsCh82:online}. Many disease areas are crowded with old and new treatments, each of which may be evidenced by several clinical trials that report varying risks and benefits in different patient groups. This makes it difficult or simply impossible for individual clinicians to keep abreast of the latest evidence through the traditional method of reading papers. Instead, they must turn to systematic literature reviews, which aggregate available evidence, primarily from RCTs, for answering predefined clinical questions and making decision recommendations. 

\begin{figure}[h]
    \centering
    \includegraphics[width=0.9\textwidth]{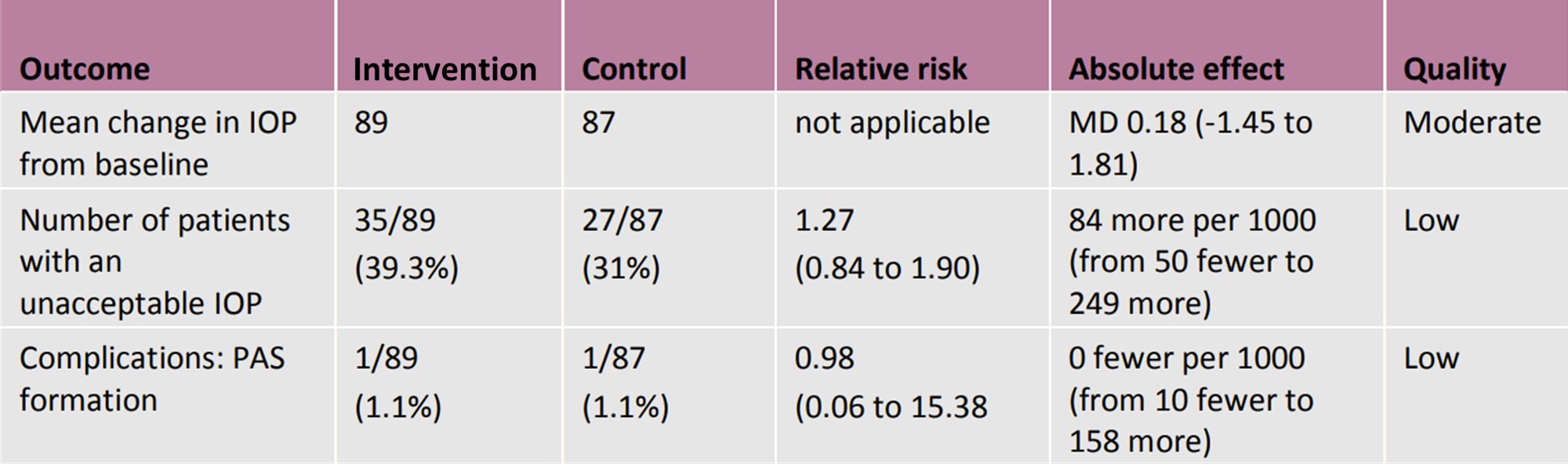}
    \caption[An example of an evidence table from the NICE clinical guidelines for glaucoma showing the results of an RCT investigating a treatment for lowering intraocular pressure (IOP).]{An example of an evidence table from the NICE clinical guidelines for glaucoma showing the results of an RCT investigating a treatment for lowering intraocular pressure (IOP). Other abbreviations: MD, Mean Deviation; PAS, Peripheral Anterior Synechiae. Figure adapted from \cite{nice_2017}.}
    \label{fig:NICE_table}
\end{figure}

Primarily conducted by healthcare bodies such as the National Institute for Health and Care Excellence (NICE) in the UK and the World Health Organisation (WHO), these reviews are labor intensive -- published RCT papers must be searched for on medical publication sites such as PubMed, Medline or UptoDate, screened for inclusion, and then read carefully to extract the relevant information into evidence tables (Figure \ref{fig:NICE_table}). To provide an appropriate summary, this information needs to be relatively extensive and detailed, including data on the trial arms, patient populations and study results. The information extraction (IE) step is generally performed using a predefined framework, which provides a consistent approach to decomposing clinical questions, so that specific and recurring data elements can be retrieved to answer them, the most common being the \textbf{P}opulation, \textbf{I}ntervention, \textbf{C}ontrol, \textbf{O}utcome (PICO) framework \cite{Chapter293:online}. 

While annotation software tools are available, much of the systematic review process still requires manual input from domain specialists that is both time-consuming and expensive  \cite{Borahe012545,michelson2019significant}. It has been estimated that the average yearly cost of systematic reviews is about 18 million dollars for each academic institution and 17 million dollars for each pharmaceutical company \cite{michelson2019significant}. An automated system that can extract information from RCTs and automatically tabulate it into evidence tables is therefore highly desirable. However, the real value of such a system lies in its potential to overcome some of the key limitations of clinical guidelines \cite{hunter2012aggregating}. New guidelines often take years to produce, and can quickly go out of date as new evidence becomes available. They also rarely account for the whole multitude of different patient characteristics and local regulations that a decision maker may face, restricted by the prohibitive cost and complexity of conducting a single or multiple systematic reviews to cover every imaginable detail. As a tool for healthcare bodies or even decision makers themselves, automated evidence aggregation could keep up with the pace of new publications and reduce the cost barrier to developing personalised recommendations for specific patient characteristics and local requirements  \cite{hunter2012aggregating}.   

There has been some headway towards this goal; a number of natural language processing (NLP) studies have looked at rules-based, statistical and, more recently, neural net (NN) models for the automated extraction of information from RCTs, achieving varied results. Many of these investigations have focused on identifying relevant information at the sentence level, requiring further methods or human intervention to process the granular detail needed for an evidence table. Extraction of more detailed information, using techniques such as named entity recognition (NER) and relation extraction (RE), has progressed greatly over the last few years, largely thanks to improvements in the way language is represented by machines, transfer learning, and the release of large gold-standard corpora of clinical trial annotations. However, to our knowledge no study to date has utilised these advances to automatically extract and differentiate intervention, outcome, and outcome measure entities into the appropriate columns of an evidence table for a systematic review.  

In the current study, we investigate automatic IE from RCTs for such a task, focusing on the tabulation of result sentences from the abstracts of published papers. To achieve this, we decompose the problem across an NLP pipeline with two transformer-based components -- an NER model for extracting entities full length entities and a RE model for identifying the relations between them -- whose output we use to construct appropriate tuples for an evidence table. Specifically, we looked to identify intervention, outcome and outcome measure entities, and use their respective relationships to sort them into tuples of the form (\textit{outcome, arm 1, arm 2}). In addition, we present a new gold standard corpus used to train our models of almost 600 result sentences from across six disease areas, with both named-entity and relational annotations. This corpus is made publicly available along with the system code in our study repository\footnote{\url{https://github.com/jetsunwhitton/RCT-ART.git}}. Evaluation of our system explores performance differences gained by fine-tuning language representation models that were pre-trained on domain-specific corpora versus fine-tuning a model that was pre-trained on a general corpus. We demonstrate that the system performs well across multiple NLP tasks and in generalising to unseen disease area domains, maintaining performance after training on as few as 200 example sentences.   

\section{Background}

\subsection{Related work}
Automating IE from healthcare publications has been the subject of considerable research, across a variety of NLP methods. PICO extraction is a key component of many of these studies, and is often a first step in more complex tasks such as argument mining and question answering \cite{stylianou2021transformed,mayer2020transformer,schmidt2020data}. However, there are significant differences across publications in terms of the PICO elements targeted for extraction, as well as their levels of detail, with investigations tending to focus on either full sentence classification or fine-grained, sub-sentence extraction of named enemies. 

\subsubsection{NER of PICO elements}
\label{sec:background_ner}

 NER involves the labelling of sub-sentence lengths of unstructured text that describe named entities with predefined categories (e.g. labelling "aspirin" as "drug"). Around a decade ago, researchers using NER to extract PICO elements generally focused on identification of only the population, intervention and control categories. These studies tended to use statistical machine learning models such as conditional random fields (CRFs), support vector machines and naïve bayes to classify noun-phrases and sentences, and then further processed these with hand-crafted rules and regular expression matching to extract PICO elements \cite{kiritchenko2010exact,hara2007extracting}. Although these approaches have been shown to be effective to varying degrees, relying on hand-crafted patterns and rules can limit a model's ability to generalise. To overcome this problem, Trenta et al. \cite{trenta2015extraction} restricted the use of rules and pattern-matching in their two-stage classification system that used a statistical ML model to identify the syntactic heads of PICO entities (e.g. ``patient" in ``patient with glaucoma" or ``corneal" in ``corneal implant") in RCT abstracts. This study included both outcomes and measure entities, and achieved encouraging results; however, classification of full-span entities was left to future research.

Another key limitation of PICO NER has been the lack of publicly available training data, with researchers such as Trenta et al. sourcing and annotating their own RCT publications. Along with presenting an expensive and time-consuming barrier to entry for this research area, the self-annotation of text creates a number of issues; the most salient being that variations in corpus annotation methods inhibits meaningful comparisons of system performance across different studies. However, steps to overcome this barrier were made in 2018, when Nye et al. \cite{nye-etal-2018-corpus} released the EBM-NLP corpus, which contains 4,993 PICO annotated abstracts of RCTs from the MedLine database, and was developed using a combination of crowdsourcing and expert review. This dataset has now been used by multiple studies to train current state of the art (SOA) models, most commonly with deep NN architectures. These include: a recurrent NN trained by Brockmeier et al. \cite{brockmeier2019improving} for identifying PICO elements to score abstract relevancy against systemic review questions; a long short-term memory (LSTM)-CRF model trained by Nye et al. \cite{nye2020trialstreamer} for the PICO extraction component in their live and automated RCT classification system, Trialstreamer; and a LSTM-CRF model by Kang et al. \cite{kang2019pretraining}, who further labelled a subset of studies from the EBM-NLP corpus with outcome measures to train one of the most comprehensive PICO NER systems to date.

\subsubsection{RE between PICO elements}

RE is a core IE task with a variety of approaches, seeking to identify the contextual relationships between sentences or entities, such as which outcome measure belongs to which study arm (e.g. identifying that the outcome measure, \textit{"39.3\% of patients had unacceptable intraocular pressure"}, belongs to the intervention arm that received the study drug, \textit{"latanoprost"}). As the gateway to semantic reasoning, identifying the relations between PICO entities is a complex task, especially if entities are jointly extracted, and is less studied than NER alone, particularly prior to the development of contextualised embeddings \cite{jonnalagadda2015automating}. Since this milestone, however, PICO extraction studies with this objective have begun to trend. 

Initially, studies of RE with contextualised language representations focused on identifying general bio-medical entity-pair relations, rather than direct PICO elements. For example, studies by Lim and Kang \cite{Lim2018ChemicalgeneRE} and Joël et al. \cite{joël2018crosscorpus} sought to jointly extract gene--disease and drug--disease entity pairs, respectively, and classify their relations, using LSTM architectures to first extract entity pairs, and then relations via a grammar dependency mechanism. More recently, Nye et al. \cite{nye2021understanding} and DeYoung et al. \cite{deyoung2021ms2} used transformer-based models to extract ICO entities (ignoring population in PICO) and their relations, using this information to construct ICO triplets, where an intervention, control and outcome description are matched with a comparative outcome description (e.g. intervention \textit{reduced} outcome compared with control).

In this study, we build on the concept of the ICO triplet, by looking beyond comparative outcome descriptions. Instead, we seek to create triplet tuples where outcome descriptions, the interventions for each study arm and their individual outcome measures are divided into respective tuple positions (i.e. columns of a table). 

\subsection{Transformer-based language representations}

Context is a fundamental component of language and presents a key challenge in the representation of language by machines, particularly in complex domains such as healthcare. However, the relatively recent breakthroughs of transformers \cite{vaswani2017attention} and transformer-based encoding architectures \cite{devlin2018bert,radford2018improving} have advanced contextualised language representations, using the mechanism of attention to embed and encode word tokens and their contextual information into a feature vector space. 

Bidirectional Encoder Representations from Transformers (BERT) are one such architecture \cite{devlin2018bert}, with its encoded representations considering contextual words from both left-to-right and right-to-left of the target word token. This is particularly important for token-level tasks, including NER and RE, where context from both directions is crucial. BERT models also employ transfer learning, where models are first pre-trained on large datasets to create reusable language representations. These representations can then be fine-tuned for a variety of downstream NLP tasks through training on relatively small task-specific datasets. The original BERT model was pre-trained on a corpus of over three billion words from the BookCorpus and English Wikipedia, which was then fine-tuned to achieve SOA performance on a large number of sentence-level and token-level tasks, even outperforming many task-specific architectures \cite{devlin2018bert}.

Extensive research has now built on the original BERT system. Some studies have adapted its architecture, such as the general models RoBERTa \cite{liu2019roberta} and ERNIE \cite{zhang2019ernie}, with the latter achieving better than human results on the SuperGlue benchmark \cite{sun2021ernie}, an aggregate of scores from a variety of NLP tasks. Others have extended BERT's pre-training with enormous domain specific corpora. SciBERT \cite{beltagy2019scibert} and BioBERT \cite{lee2020biobert} are two such systems in the biomedical domain, motivated by the belief that bidirectional context is critical for representing the complex relationships between biomedical terms. Using the weights of the original BERT model as a base, BioBERT is pre-trained on abstracts from PubMed and full text articles from the PubMed Central archive, accounting for a sum-total of 14 billion additional words, while SciBERT is pre-trained from scratch on 1.14 million papers from Semantic Scholar (over three billions words). Both models demonstrate performance gains over the original BERT model for NLP tasks in the biomedical domain. 

In the development of our NLP pipeline, we explore transfer learning with both of these domain-specific language representations, as well as the general RoBERTa model for comparison. 

\section{Dataset creation}
\label{sec:data_creation}
In this section, we discuss the development of our dataset used to train and test our RCT result tabulation system. The novel data included in our corpus was created through annotating sentences from structured RCT abstracts. We collected these abstracts from two sources, the Trenta et al. \cite{trenta2015extraction} study dataset of glaucoma RCT abstracts and the EBM-NLP corpus \cite{nye-etal-2018-corpus}, which were preprocessed before either reannotation (Trenta et al. dataset) or additional annotation (EBM-NLP corpus) for the given task at hand. Our final gold-standard dataset included over 595 annotated sentences from six disease-area domains.   

\subsection{Data collection}
As the primary evidence in EBM systematic reviews, we used RCTs as the source of data for our study. In particular, we decided to focus on the abstracts of published RCT papers, as these are both freely available and offer a structured summary of the trial, which should conform to the CONSORT policies published in 2010 \cite{moher2012consort}. These guidelines outline how RCT publications should be constructed, and include a checklist to ensure all key trial information and results are reported within the abstract, within the following labelled sections: background, objective, method, results and conclusion. In addition to ensuring they are a reliable RCT overview, these minimum requirements for information mean abstracts from different studies are comparable in terms of included PICO entities, even across different disease-area domains.  

In line with our focus of extracting respective relationships between study arms, outcomes and measures, we made two further key restrictions to data collection, adapted from those used by Trenta et al \cite{trenta2015extraction}. First, we only included abstracts from RCTs with a two-group study design. This was done to simplify our task and limit the study's scope, as sentence complexity of reported results, as well as the number of entities and relationships to track, increases in line with the addition of study arms beyond two. Second, we limited our dataset to abstracts with a result sentence that includes at least one study arm, outcome and a clear, numerical measure with respective relationships to each other, examples of which can be seen in \autoref{sec:anno}. These result sentences were annotated to form the novel training and test data used for framing the NER and RE tasks as supervised-learning classification problems.

\subsection{Abstract sources}

\subsubsection{Trenta et al. glaucoma}
Our dataset was initially composed from that of the Trent et al. \cite{trenta2015extraction} study, which similarly investigated the extraction of PICO information with respect to study arms, albeit with a different approach that focused on statistical-based methods. This dataset comprises RCT study abstracts in the disease area of glaucoma, collated from PubMed with three search strategies: 1) titles and abstracts including ``glaucoma" and specifying the study as an RCT; 2) titles including at least one prescription drug for glaucoma or ocular hypertension from a predefined list, and specifying the study as an RCT; 3) titles including at least one surgical procedure for glaucoma or ocular hypertension from a predefined list, and specifying the study as an RCT. These queries retrieved 176 abstracts, with this set filtered using inclusion criteria similar to those outlined in the previous section. We have included all 99 of the resulting abstracts in our dataset, with 214 result sentences included in our training and testing corpus.

The annotation approach of the Trenta et al. was to label the syntactic head of each target element within an abstract, including population, study arm one, study arm two, the outcome description and the respective outcome measures for each study arm, with the limitation of labelling just one of these elements per abstract. However, with the BERT-based approach of our study, a new annotation schema, described in the next section, was necessary to optimise the learning of the system. 

\subsubsection{The EBM-NLP corpus}

To improve the generalisation capacity of our trained system, we looked to extend our dataset beyond one disease-area domain with the recent EBM-NLP corpus \cite{nye-etal-2018-corpus} for PICO extraction. The corpus is comprised of 4,993 RCT abstracts sourced from the MedLine with a general focus on the domain areas of cardiovascular disease, autism and cancer, covering a range of common conditions. Unlike the glaucoma dataset of Trenta et al. \cite{trenta2015extraction}, the EBM-NLP corpus was developed with no particular data extraction method in mind, and as such, had no inclusion criteria beyond requiring abstracts to describe an RCT. Because of this, and the lack of separation between domain data, it was necessary to screen the data for collection from this corpus, which we accomplished through a mixture of automated techniques and manual screening at the annotation stage. We retrieved datasets for five disease-area domains, including autism (208 abstracts), blood cancer (74 abstracts), solid tumour cancer (200 abstracts), diabetes (97 abstracts) and cardiovascular disease (159 abstracts), from which we included 47, 16, 130, 51 and 130 sentences in our train and test sets, respectively.   

There are two datasets within the EBM-NLP corpus, comprised of the same abstracts but differing in annotation detail: the first includes broad labels for population, intervention (covering both interventions and controls) and outcome entities, while the second includes fine labels highlighting more detailed information, such as sample size, patient characteristics and outcomes types (adverse events, survival, treatment impact, etc.). However, neither of the two datasets label numeric measures of outcomes. To restrict the complexity of our system's classification task, we selected the dataset with the first of these annotation types, and extended them with measure labels in line with our study objective.

\subsection{Annotation}
\label{sec:anno}
Our gold-standard data was created with a two-stage annotation process, entity labelling and relationship labelling, using the Prodigy python package by Explosion \cite{Prodigy71:online}. Prodigy is an industry-grade annotation tool for efficient, local-server, browser-based text labelling across a variety of NLP tasks, and was made available to us through a free research licence. As highlighted in the data collection section, only result sentences with at least one numerical measure of an outcome with respect to a treatment arm were included for novel annotations (extending the pre-existing annotations of the NLP-EBM dataset). Comparative measures (e.g. an odds ratio comparison of two risk measures) are omitted.   

\subsubsection{Entity annotation}
\begin{figure}[h]
    \centering
    \includegraphics[scale=0.4]{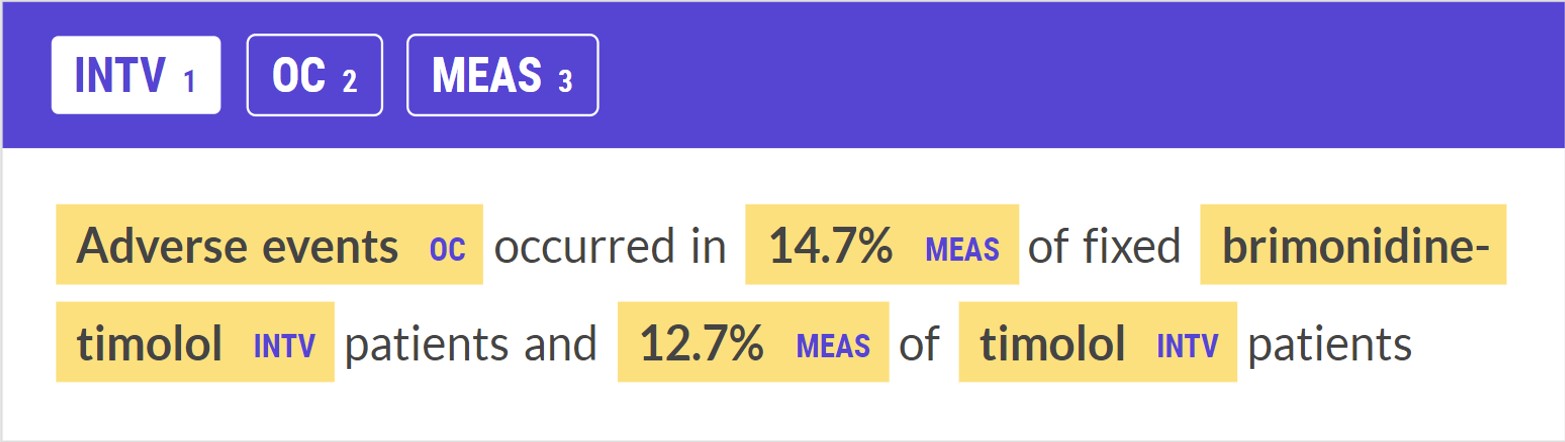}
    \caption{NER annotations in Prodigy.}
    \label{fig:ner_annotation}
\end{figure}

Three types of entity (Figure \ref{fig:ner_annotation}) were labelled during this stage of the annotation process: interventions (INTV), outcomes (OC) and  measures (MEAS). The annotation guidance for each of these labels is as follows:

\begin{itemize}
 \item \textbf{INTV:} label the study treatment or its comparator that patients have been randomised to receive
 \item \textbf{OC:} label any description of the study results, most commonly describing measures of effectiveness and safety
  \item \textbf{MEAS:} label any clear, non-comparative, numeric measure that can be related back to an outcome and/or intervention within the sentence 
\end{itemize}

 A decision was made early in the study not to distinguish between the entity labels of interventions and their comparators, as well as their respective outcome measures (i.e. separate labels for the intervention measure and comparator measure). While identifying these entities would make the task of dividing data into the study-arm columns of an evidence table relatively trivial, the similarly fine-detailed labels of the EBM-NLP corpus were shown to significantly reduce model performance ($F_1$ score reduction of from 0.68 to 0.46 vs broad labels) \cite{nye-etal-2018-corpus}. Moreover, outcome measures need to be related back to their respective outcomes, which would necessitate either even more specific entity labels or overlapping entity spans, the latter being highly non-trivial for NER models \cite{muis2018labeling}. Instead, we sought to decompose the complexity of the task by sub-dividing it between the NER and RE components of our system. Masking the intervention--comparator division allows our NER component to focus on classifying a simpler set of entity labels, which are then passed to our RE component, where entities are further distinguished through classifying the relations between entity pairs.

\subsubsection{Relationship annotation}
\begin{figure}[h]
    \makebox[\textwidth][c]{\includegraphics[width=1.05\textwidth]{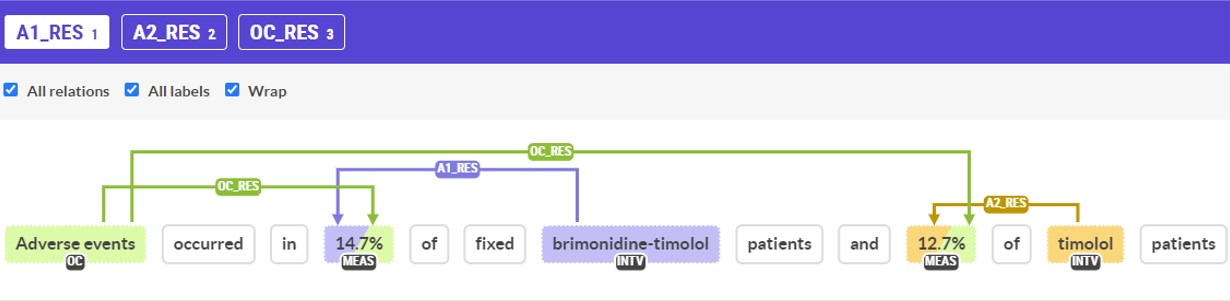}}
    \caption{RE annotations in Prodigy.}
    \label{fig:re_annotation}
\end{figure}

Entity relations were labelled between relevant entity pairs (Figure \ref{fig:re_annotation}) in each sentence, with three types of relationship labels: relations between measures and their \textbf{RES}pective outcome description (OC\_\textbf{RES}), and relations between a measures their \textbf{RES}pective intervention arms, which we have limited to two in the current investigation (A1\_\textbf{RES}, A2\_\textbf{RES}). Relationship labels were also directional, highlighting a parent-to-child dependency, with the INTV and OC being parent entities and MEAS being the child entity. 

\begin{figure}[h]
    \centering
    \includegraphics[scale=0.5]{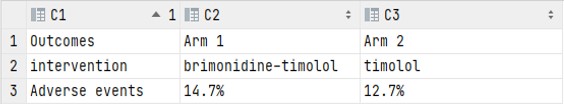}
    \caption{Output CSV of a gold-standard table, generated by our tabulation component from the gold-standard NER and RE annotations.}
    \label{fig:gold_table}
\end{figure}

Although further relations could have been annotated, such as the comparative relationship between study arms, we decided to limit complexity through restricting these labels to the minimum needed for our tabulation component to construct a result sentence into an evidence table (Figure \ref{fig:gold_table}), with a column for each arm and row for an outcome description. 

\section{System design and implementation}
\label{ch:implementation}

We discuss the implementation and design of our NLP pipeline for RCT result sentence tabulation in this section. Our system was designed to house two key components -- a model for NER of INTV, OC and MEAS elements and a model for extracting the relations between them  -- both built on contextual BERT-based language representations and transfer learning. The output of these models is then processed by a tabulation module, which returns result sentence tables in CSV format. In addition, we present Python functions and classes used in data collection from the EBM-NLP corpus, preprocessing and for adapting inbuilt spaCy components for our pipeline. 

\subsection{System architecture design and overview}

Our  extraction system was developed in Python 3.9.6 with the open-source NLP library spaCy (version 3.1). spaCy includes a variety of NLP tools for tasks ranging from rule-based sentence segmentation to BERT-based NER and was developed for building custom language processing pipelines \cite{neumann2019scispacy, le2021taxonerd}. It also connects to the HuggingFace transformer library, which allowed us to import different BERT-based language representations for use in our models, including SciBERT, BioBERT and RoBERTa.

\begin{figure}
    \centering
    \includegraphics[scale=0.9]{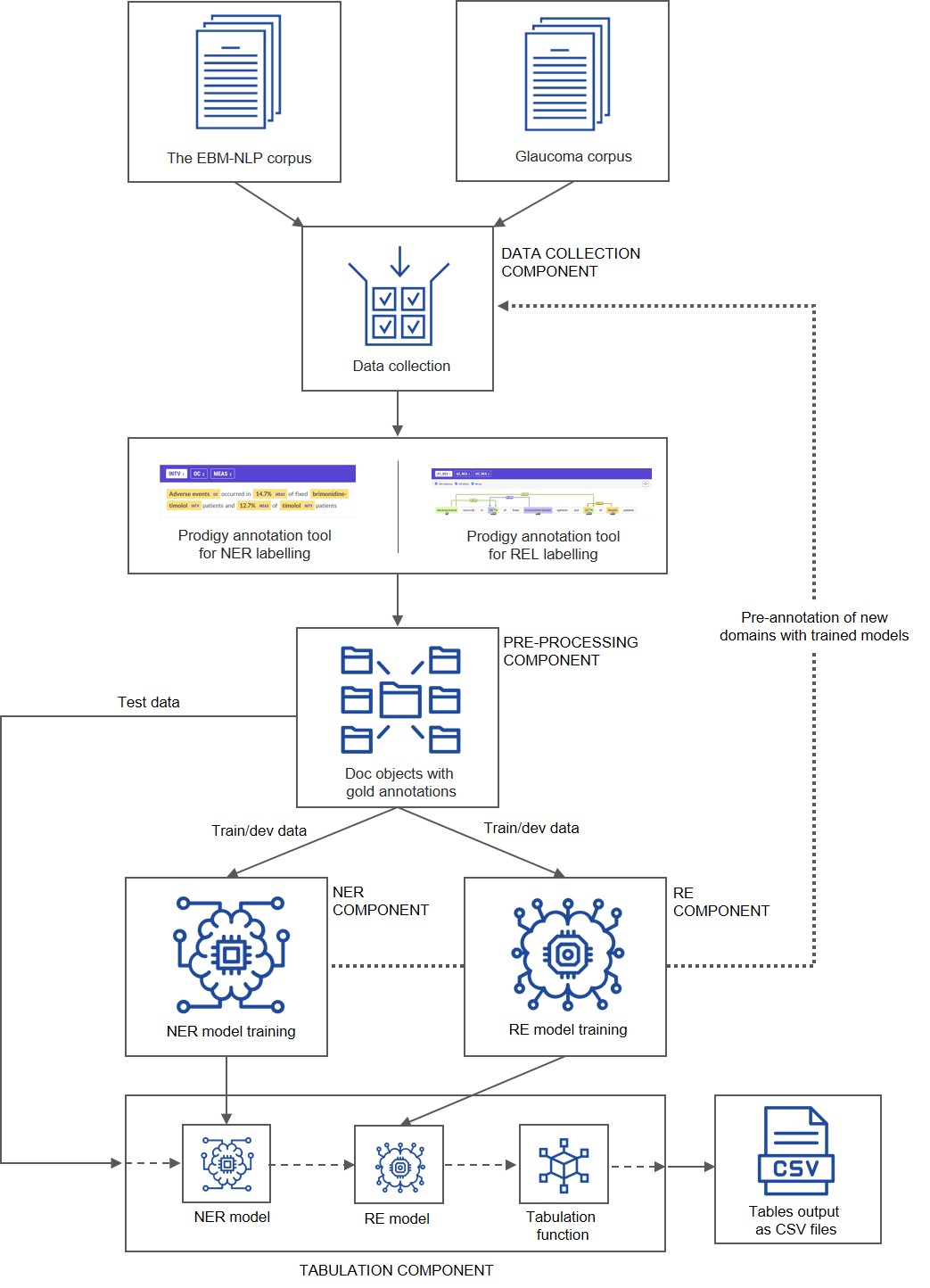}
    \caption{The architecture of our full study system.}
    \label{fig:architecture}
\end{figure}

An overview of the full architecture of our study system can be seen in Figure \ref{fig:architecture}, and consists of five key components. The first is a data collection module for retrieving abstracts from the EBM-NLP corpus and Trenta et al. study dataset, and processing them for annotation with Prodigy. The second component is a preprocessing module that prepares the annotated data for training our spaCy pipeline models. Performing the key tasks of our system, these models form our third and fourth components and are for NER and RE, respectively. The final, fifth component tabulates result sentences, applying both the NER and RE models sequentially to input text, outputting these tables as CSV files. 

\subsection{The data collection component}
\label{sec:data_collection}

Built as a Python module, our data collection component has two parts: a class for retrieving and processing abstracts from the EBM-NLP corpus; and a collection of functions for processing retrieved abstracts before annotation with Prodigy.

\subsubsection{Class for EBM-NLP dataset processing}

The EBM-NLP corpus datasets are made available as a collection of abstract texts and associated token annotations in a format used by the web-based annotation tool brat \cite{devlin2018bert}, which is unsupported by the spaCy and Prodigy libraries. Furthermore, they are unsorted in terms of disease area, which we required for testing the generalisation ability of our system across domains. To overcome these issues, we developed a class for processing the EBM-NLP broad-labelled dataset, which includes a pipeline of functions for query-based retrieval of abstracts, and processing these into a format supported by the spaCy and Prodigy libraries.

A brief overview of the retrieval pipeline starts with the PubMed IDs (PMIDs) naming system of the abstract text files, which are extracted for querying the PubMed database using the BioPython library and Entrez. To overcome querying limitations of the database, the 5,000 PMIDs of the dataset are batched into ten lists, with each list concatenated to our predefined terms (e.g. the disease-area ``diabetes"), with construction used to retrieve the PMIDs of abstracts matches for our predefined term. This querying approach to a PMID ``filter" allowed us to quickly sort the abstracts of the NLP-EBM corpus into disease areas for cross-domain testing. 

Once identified, abstracts from each disease area and their annotations in the brat format are converted into the spaCy Doc format, the primary data-structure used by our model, which includes the full text, as well as its tokens and labelled entities. This process first requires converting them to an IOB format and updating the entity tags with our study labels (INTV, OC, MEAS).

\subsubsection{Functions for pre-annotation processing}

The functions outside of the EBM-NLP class within the data collection module are geared towards selecting and preparing data for novel annotations.

Abstracts texts are first segmented into sentences, as long texts require greater resources for ML models to process, with time complexity increasing quadratically with sequence length ($O(L^2$); $L$ = length) for BERT-based models \cite{zhang2019bert}. Sentences were chosen as the minimum sequence length to retain enough information to map contextual relationships between entities. In line with our data collection strategy, result sentences are then retrieved to form the annotation dataset, with their source abstracts kept track of with a PMID attribute.

The last function in this module was for pre-annotation. As domains were annotated sequentially, we trained our models on the gold-standard labels as they became available, and used their predictions to pre-label subsequent domain datasets. This allowed our annotators to correct labels rather than annotate from scratch, greatly improving efficiency. 

The final output of these functions is in a JSONL format used by the Prodigy annotation tool, a txt-type file where each line is a JSON object including the sentence text, its tokens and its pre-annotation labels (i.e. entities and the relations between entity pairs). 

\subsection{The preprocessing component}

Similarly to the data collection component, the preprocessing component was built as a Python module, and was used to process our gold-standard datasets after annotation into training and test input for our models.

Fine-tuning the contextualised language representation of BERT, trained on billions of words from unprocessed text, does not require input text to be preprocessed with normalisation techniques such as lemmatisation or stemming, and can benefit from keeping stop words, which form important parts of word contexts. We considered normalising medical abbreviations (e.g. replacing IOP with \textbf{i}nter\textbf{o}cular \textbf{p}ressure); however, as our key base models for transfer learning, BioBERT and SciBERT, were also trained on millions of unprocessed medical abstracts and papers, where abbreviations occur frequently, we made the assumption our model would be able to perform well in classifying these tokens. Therefore, because tokenisation and sentencisation are both performed, by necessity, during the data collection stage, a limited amount of further preprocessing was required for our gold data, and was achieved with a small number of functions. 

First, sentences rejected during the annotation process are screened and removed from the dataset. After filtering, the gold-standard annotations were merged into an all-domains dataset, as well as stratified with varying numbers of domains, all used to evaluate the model's performance for each disease area and ability to generalise across them. The merged and individual domain datasets were then converted from the Prodigy JSONL output to spaCy Doc format, with a function that extends them with a custom attribute for the entity pair relation annotations. Before training our model, all datasets were split into a training (train), development (dev; also known as validation set) and test set in a 70/10/20 split, with sentence order randomised before the split. The former sets were used to train our models, and the latter unseen data used to test them and the full system. 

Lastly, a number of helper functions were developed for splitting our datasets into different sizes and domain mixtures, for training and testing models in different situations as part of the evaluation phase of the study. 

\subsection{The NER component}
Our NER component was developed as a spaCy pipeline, with a language representation model feeding into the NER prediction model provided by the library. Importantly, this NER extractor works with interchangeable language representations, including pre-trained transformer-based models.

\subsubsection{Design}
The spaCy NER model has a transition-based parser architecture, inspired by the chunking model from Lample et al. \cite{DBLP:journals/corr/LampleBSKD16}, where a sequence of tokens is incrementally passed from a buffer to an entity stack for labelling or straight to the output list, with a greedy algorithm choosing the optimal action to take. The spaCy adaptation of this architecture has five actions to choose from:

\begin{itemize}
    \item[] \textbf{Begin:} Begin a new entity by adding token to stack
    \item[] \textbf{In:} Continue the current entity by adding token to stack
    \item[] \textbf{Last:} Entity label stack with current token as last word, move stack to output
    \item[] \textbf{Unit:} Label current token as a single-word entity and move to output
    \item[] \textbf{Out:} Move current token straight to output without marking as entity
\end{itemize}

\begin{figure}[h]
    \centering
    \includegraphics[scale=0.4]{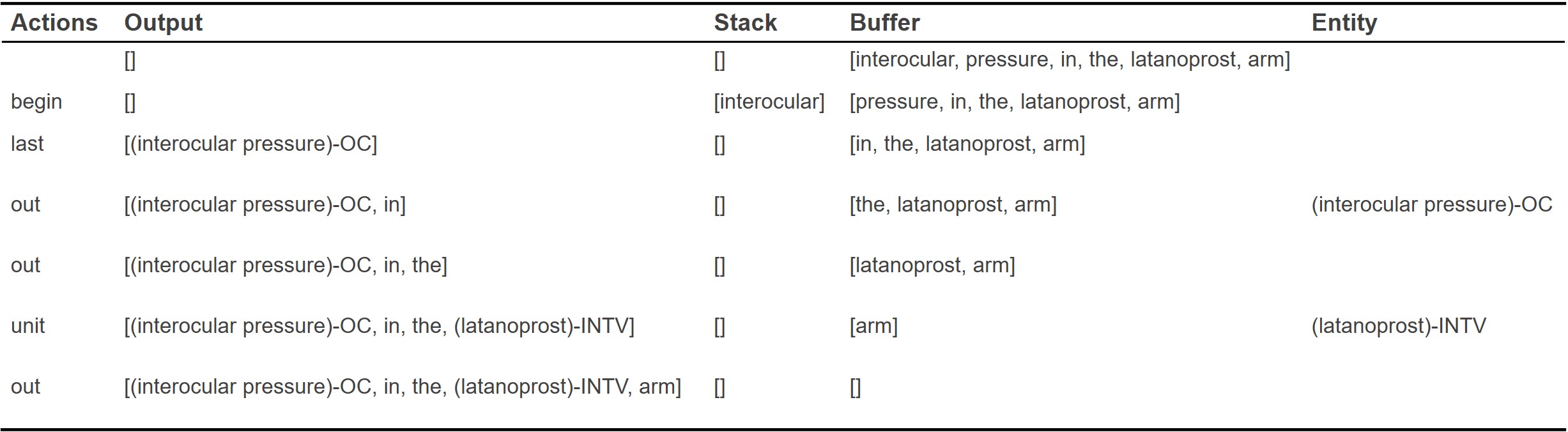}
    \caption{Example of optimal actions taken as an input sequence is passed into the dependency parser.}
    \label{fig:dependency}
\end{figure}

We describe an optimal set of actions with an example sentence from the glaucoma domain in Figure \ref{fig:dependency}. The scores for each possible action are calculated at each time step by feeding a representation of the current state of the stack and buffer to a multilayer perceptron, with the best action chosen until the algorithm reaches a termination state. The state representation is derived through combining the word embeddings of the tokens that make up the entity stack and the buffer, which are passed to the transition-based parser layer from an upstream language representation model, in our case a pre-trained transformer model. 

\begin{figure}
    \centering
    \includegraphics[scale=0.50]{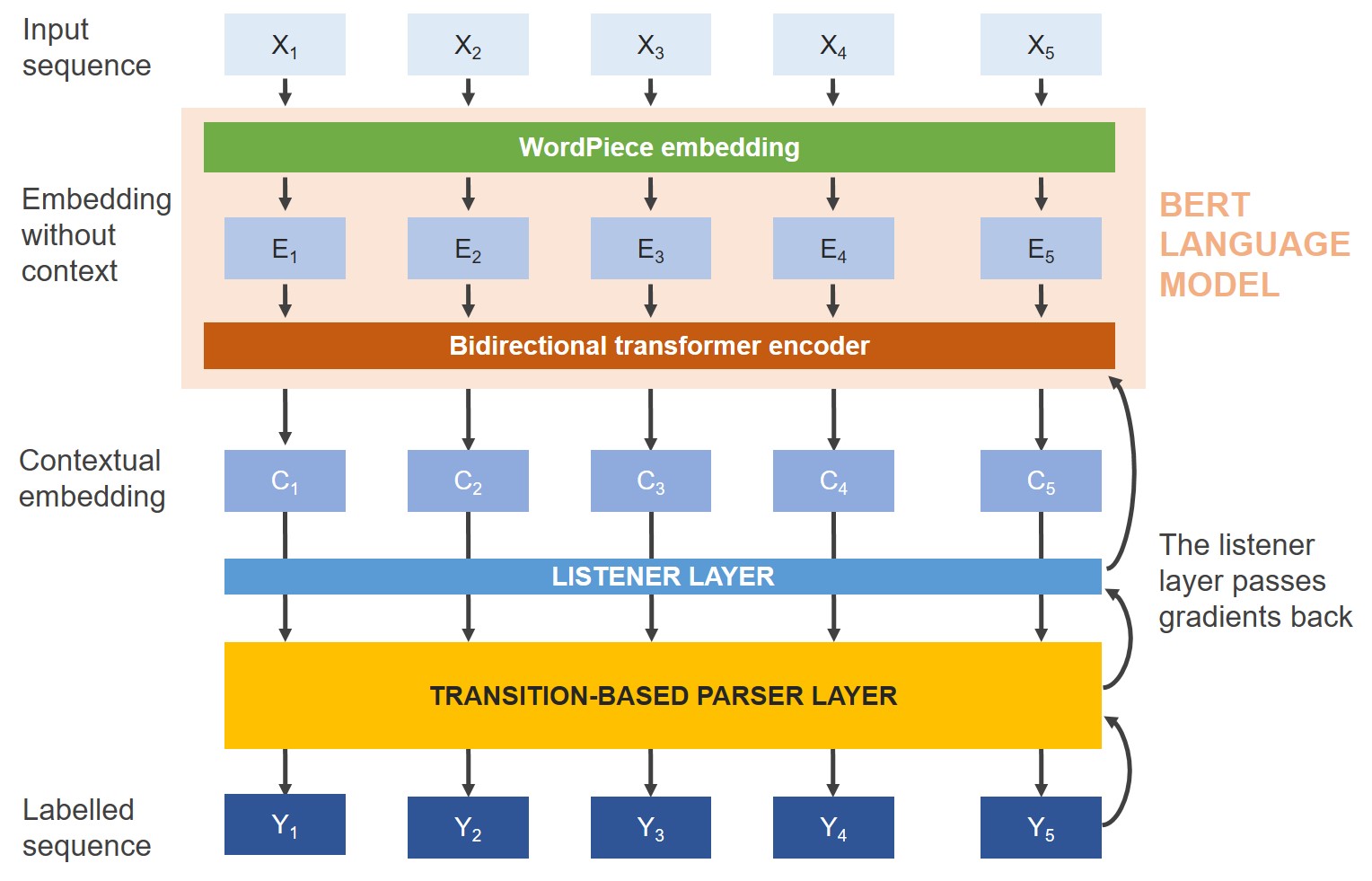}
    \caption{Schematic of the NER component architecture.}
    \label{fig:ner_arch}
\end{figure}

Components in a spaCy pipeline are built with ``listener" layers that allow them to receive word embeddings from layers made of interchangeable language models. These models embed and encode tokenised input sequences (study abstract sentences in our study) in the Doc format and output tensors of predicted word vectors, passed by the listener downstream to the NER layer for calculating action scores. Our architecture leverages transfer learning by including a pre-trained BERT model as the embed and encode layer. An overview of the layers of this model design is outlined in Figure \ref{fig:ner_arch}. The listener layer of the NER model is also used for back-propagation, passing the error gradients used to adjust the model weights back upstream to fine-tune the language representation layers. Gradients are calculated from a loss function that scores entity prediction errors per action against the target gold annotations, with all layers of the NER component being updated during the target task.

\subsubsection{Implementation}

The pipeline for our NER model was implemented as a spaCy config file, a feature of the library that facilitates and centralises pipeline design and modularity, where each layer of the model is defined, along with their hyper-parameters for training. There are a large number of variables that can be adjusted within the spaCy config system for customising models. Here, we describe the main config choices made in the implementation of our system, and refer the reader to the spaCy library documentation for settings not discussed \cite{LibraryA39:online}.

For our transformer-based language representation layers, we experiment with three pre-trained BERT models: SciBERT and BioBERT, both extensively trained on medical study abstracts, and the generally trained model Roberta, which was included for comparison. We chose the ``cased" version of all these models, where the capitalisation of words is considered, as these have been shown to perform better for NER, with evidence for this in the medical domain \cite{abadeer-2020-assessment}. A separate config was created for each pre-trained model. All three models are available through the HuggingFace model database, and are defined within the transformer model section of the config file as pip install links. At training, the system initialises the pre-trained weights of these models if they are already installed in the local environment or downloads them from the database for installation and initialisation. 

Our hyper-parameters for training were initially set to match those used by Devlin et al. \cite{devlin2018bert} for fine-tuning the original BERT model. These included a training example batch size of 32, a dropout rate of 0.1 (a regularisation parameter where a proportion of neurons are randomly ``dropped" from the network), and a learning rate of 5e-5 with Adam optimisation used to fine-tune the system. After experimenting with a limited number of alternative hyper-parameter values (batches: 16, 32, 64, 128; drops: 0.1, 0.2, 0.3; learning rates: 5e-5, 3e-5, 2e-5), in our final model configuration we adjusted the batch size to 64 and the dropout rate to 0.2, both gaining us incremental performance increases on our test data. 

Models were trained with early stopping, activated by a patience parameter of 1,000 steps with no performance gains on the development set, with a limit of 20,000 steps and no cap on epochs (the spaCy framework is uncommon in using steps rather than epochs for patience). A single machine was used to train all models, operating with a GeForce RTX 3080 GPU (10GB video RAM) and 16GB of RAM.

\subsection{The RE component}

Similar to our named-entity extractor, the RE component of our system was built as a spaCy pipeline, again using BERT-based language representations. However, as the library does not include an inbuilt RE model, we needed to implement it as a custom component of the pipeline, which we adapted from a spaCy project template \cite{projects91:online}.

\subsubsection{Design}

\begin{figure}
    \centering
    \includegraphics[scale=0.5]{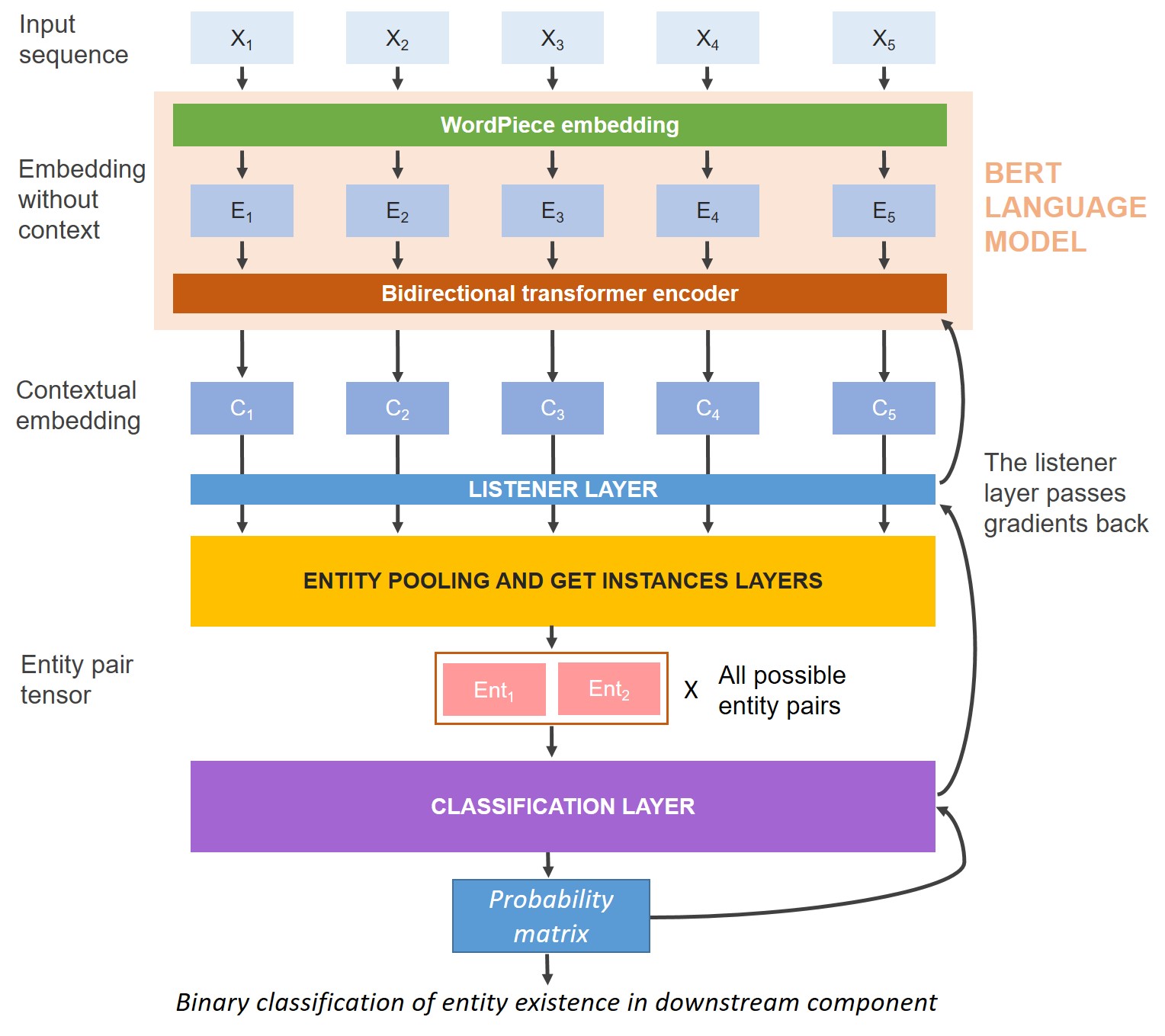}
    \caption{Schematic of the RE component architecture.}
    \label{fig:re_arch}
\end{figure}

Designed with multiple layers, our RE model is built with a multi-label classification objective, where it scores a probability for each of our relation labels between entity pairs within the input sequence. As with our NER model, a listener layer passes output tensors of predicted word vectors downstream, calculated by a pre-trained transformer model from our tokenised input sentences. In this case, however, two extra layers are required before classification. The first extracts word vectors for entities (checking the entity span labels of the Doc object), with the vectors of multi-token entities being ``pooled" into a single vector, by taking their mean. The second pairs potential relation instances of entity pairs, in both directions to assess parent (i.e. subject) and child (i.e. object) status, outputting these instances as a tensor of the paired entity vectors. For classification, the output tensors are forwarded to a linear layer, where the vectors are multiplied by a weights matrix and a bias vector is added, and then to a sigmoid activation function for multi-label classification. The final output is a probability matrix for each entity pair across all of the defined relation labels, and both possible parent--child directions of the pair. A full overview of this architecture can be reviewed in Figure \ref{fig:re_arch}. 

The relation-type label of entity pairs is identified by the highest probability in the matrix, with a hyper-parameter probability threshold value being set for the existence of any relation at all (e.g. if no probability in the matrix is above $P(0.5)$ then no relation is classed). This binary classification is performed on the matrix output of the model architecture above by a downstream component (see next section) or at system evaluation.

For training, the loss function is calculated as mean square error, taking the difference between predicted probabilities of the output matrix and gold standard annotations, where probabilities are set to one for existing relationships and zero for non-existing relations. Again, back-propagation is used to train each layer of the pipeline, with the listener layer passing gradients back upstream to fine-tune the transformer language representation. 

\subsubsection{Implementation}

Two Python modules underlie the RE component architecture outlined in the previous section: a module for the functional layers of the RE model, built using the Thinc library (an ML library cousin of spaCy, built on Pytorch), and a pipe module that integrates these layers for use in a spaCy pipeline. Functions within these modules are made accessible as model layers to the spaCy config system through @ decoration, which as with the NER component, was used to define and train the RE component. 

We again created three separate configs for testing different language representation layers -- one for each of the same pre-trained BERT models used in the NER component. For the layers of the RE models itself, we experimented with different criteria for selecting an entity pair instance for classification, first looking at a more restrictive approach that only selected pairs of interest ([OC and MEAS] or [INTV and MEAS]). However, this was found to reduce model recall performance by around 10\% without improving precision. As a result of this exploration, the final model had a relaxed function for entity pair instances, retrieving all possible combinations of separate entities within a certain distance of 100 tokens of each other.

Hyper-parameter tuning was explored with the methodology we used for the NER component, which resulted in the selection of the same parameters for training (batch size: 64; dropout rate: 0.2; learning rate: 5e-5 [Adam Optimisation]). The RE models were also trained using early stopping with the same step and epoch parameters, and on the same machine.

\subsection{The tabulation component}

The final component structures the full pipeline for the task of tabulating result sentences, using the joint predictions of our NER and RE components. Developed as a Python module, it processes batches of input sentences in Doc format over three stages. The initial two stages load the trained models to sequentially extract the entities and relations of the full input batch, adding these predictions to the Docs. These are then forwarded to the tabulation function, where the entity and relation labels are used to parse the sentence content into a structured table, which is outputted as a CSV file.

This latter tabulation function is where the probability threshold hyper-parameter for a detected entity--pair relation is set. As would be expected, a lower threshold improves recall performance, while a higher one improves precision. We found that the optimum threshold as measured by $F_1$ score was 0.5, which was selected for the final system. 

\section{Evaluation}

In this section, we report the results of our system evaluation across five IE tasks: NER, RE on gold-annotated entities, joint NER + RE, tabulation with exact tuple matching and tabulation with relaxed tuple matching. Test-data performance with different BERT-based language representations is assessed for the full dataset across all-domains. We also investigated our system's ability to execute these tasks when trained with a varying number of examples and at generalising to unseen disease area domains. The section is closed with an error analysis of the system models.

\subsection{Evaluation methodology}

The system was evaluated in terms of precision ($P$), recall ($R$) and their harmonic mean, $F$-score -- the latter being calculated with no emphasis on either of the former ($F_1$ score). All three metrics are calculated by comparing predicted to gold standard labels, as functions of the number of true positives ($tp$; correct prediction made), false positives ($fp$; incorrect prediction made) and false negatives ($fn$; no prediction made for existing gold label). We express these functions formally below: 

\begin{equation}
P = \frac{t p}{t p+f p}
\end{equation}
\begin{equation}
R = \frac{t p}{t p+f n}
\end{equation}
\begin{equation}
F_1 = \frac{t p}{t p+\frac{1}{2}(f p+f n)}
\end{equation}
\\\\
For our multi-label classification tasks (NER, RE and joint NER + RE), due to our labels being slightly imbalanced (all-domains NER labels: 1,163 INTV, 1,526 MEAS, 820 OC; all-domains RE labels: 747 A1\_RES, 660 A2\_RES, 1,468 OC\_RES) and no priority differences between classes, we used micro-averaging to obtain $F_1$ scores, with $tp$, $fp$ and $fn$ being summed globally across classes. Only NER could be evaluated using inbuilt spaCy library tools, which meant custom evaluation functions had to be developed for assessing the other system tasks. Here, we give a brief outline of how each of these was measured.

\subsubsection{NER evaluation}
The inbuilt spaCy NER evaluation function assesses $tp$, $fp$ and $fn$ on a per-entity basis with exact match. That is, it does not count partial matches of multi-token entities as $tp$. For the NER task scores, all of the exactly matched entities across classes are marked as $tp$, with the set of predicted not in gold counted as $fp$, and the set of gold not in predicted counted as $fn$. 

\subsubsection{RE evaluation on gold entities}
To investigate the performance of its underlying model, the RE task was assessed on test data with gold-standard entity annotations. Predicted entity-pair relation labels (above probability threshold of 0.5) are counted as $tp$ if they matched the gold relation label of the same entity pair tuple (ordered tuples: $(a,b)$ and $(b,a)$ represent different parent--child relationships). Predicted class labels not matching the class of the reference annotation are marked as $fp$. Those that do not breach the classification threshold, but were in the list of gold relation annotations are marked $fn$. 

\subsubsection{Joint NER + RE evaluation}
The joint NER + RE task involves first predicting named-entities within an input sequence and then the relations between these predicted entity pairs. Evaluation of this task was somewhat more complex than the prior two, and was achieved through extending the RE evaluation function. First, all entity pairs are checked to see if they have relation annotations within the gold dataset. Those that do, are passed through to the RE evaluation function and assessed in the same way as previously described. Entity pairs not within the gold dataset are checked to see if they have relations above the prediction threshold, and classed as $fp$ if they do. Predicted entities with no relation and not part of a gold annotated entity-pair (not all gold-labelled entities have relations) are evaluated with the prior NER evaluation methodology described. 

\subsubsection{Tabulation strict tuple matching}
To assess the overall performance of the full system in tabulating RCT result sentence, tuples (order: \textit{outcome}, \textit{arm 1}, \textit{arm 2}) from the predicted output CSV files were matched against tuples from corresponding (same input sentence) gold-standard CSV files, with two matching criteria. The strict criteria required for tuples to exactly match, both in order and entries, for the prediction to be marked as $tp$. Predicted tables without an exact match were counted as $fp$. The system outputs a CSV for every input sentence, so the number of predicted and gold CSVs always match. Therefore, if a predicted output CSV is empty, it is counted as $fn$.  

\subsubsection{Tabulation relaxed tuple matching}
The second tuple-matching criterion was implement after inspection of output CSVs found that, while not matching exactly, many predicted entities overlapped with gold CSV entities (see \autoref{sec:error}). Inspired by the Type Matching criterion used by a number of biomedical entity studies \cite{segura2013semeval,le2021taxonerd, li2020survey}, relaxed tuple matching marks a predicted tuple as $tp$ if the entities have some token overlap with those of the corresponding gold tuple, and match its order. 

\subsection{System performance on the all-domain dataset with different language representations}

Results across the five tasks for each BERT-based version of our system, trained and tested on the all-domains dataset, can be found in Table \ref{table:all_results}, and are given in terms of $P$, $R$ and $F_1$ scores.  

The highest $F_1$ scores for the independent tasks of NER and RE on gold entities were 0.79 (SciBERT) and 0.77 (SciBERT), respectively. For the dependent tasks with strict entity matching, highest scores were 0.69 (SciBERT) for joint NER + RE and 0.58 (BioBERT) for strict tabulation. As these latter two tasks are downstream in the pipeline, the first dependent on the NER component and the second dependent on joint NER + RE, their respective performance decreases are to be expected. The highest $F_1$ score for relaxed tabulation, however, was 0.78 (SciBERT), likely reflecting the impact of loosening performance dependency on the upstream task of NER.

\begin{sidewaystable}
    \begin{tikzpicture}
        \includegraphics[scale=0.56]{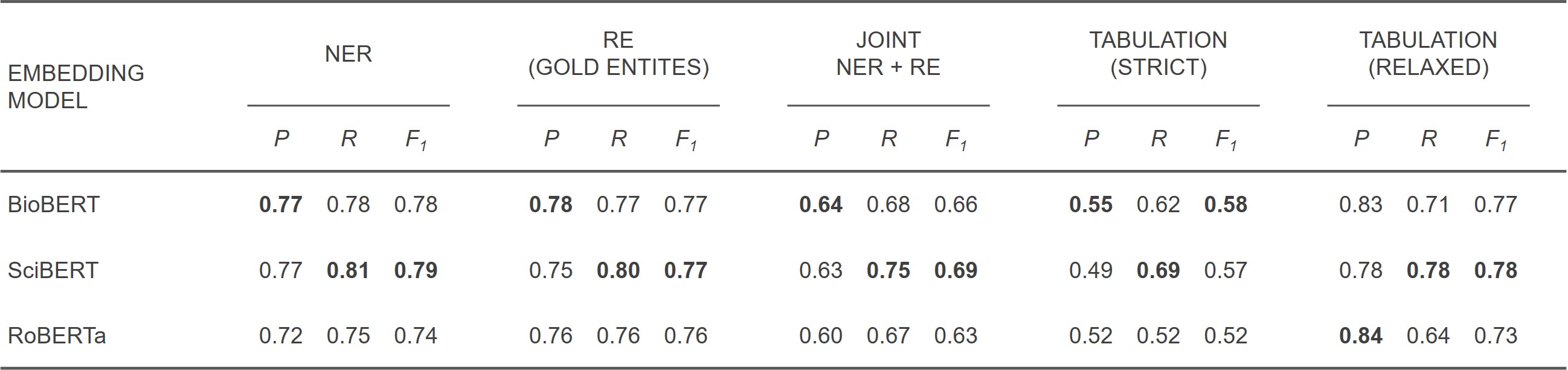}
    \end{tikzpicture}
    \caption{System performance results of the five IE tasks for each BERT-based language representation trained on the all-domains test set. Values in bold indicate the language model that scored the highest on a task-specific metric.}
    \label{table:all_results}
\end{sidewaystable}

Comparing each language representation across the five IE tasks, we found that transfer learning from models pre-trained on medical literature (BioBERT and SciBERT) performed better than from the generally trained model (RoBERTa), with \textit{$F_1$} scores around 0.05 points higher on all tasks apart from RE on the gold entity set, where the difference was 0.01 points. This follows expectations set by the literature, where domain-specific BERT models tend to perform better on in-domain tasks \cite{beltagy2019scibert,lee2020biobert}; however, the RE result is interesting. We hypothesise that it may be due to RE placing a greater emphasis on the contextual words around named entities, such as verbs and pre-positions, which hold syntactic information such as subject--object ownership (e.g. [INTV] \textit{achieved} [OC] \textit{of} [MEAS]), and are generally not domain-specific. Therefore, when given gold-standard entities, RoBERTa is able to perform similarly to its domain-specific counterparts. This may be further evidenced by RoBERTa achieving the highest precision score (0.84) on the relaxed tabulation task, which has less restrictive NER measures, but still the same RE-dependent requirement of correct tuple order.

The performance differences between the two domain-specific language models were less pronounced, with \textit{$F_1$} scores within 0.01 points of each other on all tasks other than joint NER + RE, where the difference was 0.03 points in favour of SciBERT. BioBERT tended to perform better in terms of precision, whereas SciBERT had better recall scores. This variation in performance metric ability could account for why SciBERT was better at joint extraction; as our RE component classifies relation probabilities between all available entity pairs, downstream recall losses are alleviated when more pairs are passed from the NER component.   

\subsubsection{NER performance on individual class labels}
\label{subsec:ner_perform}

System performance at classifying the individual NER class labels of the all-domains test set can be viewed in Table \ref{table:ner_labels}. 

The MEAS label was the highest performing classification with an $F_1$ score of 0.83 (BioBERT), followed by INTV with 0.79 (SciBERT) and, lastly, OC with 0.74 (BioBERT). These differences are similar to those of the literature, with outcomes tending to be one of the harder PICO elements to classify, most likely due to high variation in entity length and poor inter-annotator agreement on their boundaries \cite{nye-etal-2018-corpus}. Our best-performing NER results for measurement labels are higher than those of the Kang et al. \cite{kang2019pretraining} study; however, our dataset was more restricted in focusing solely on result sentences, rather than full abstracts. These high scores may be due to  
 measurements in clinical trials varying primarily in their parameters and values, allowing distinct lexical patterns to be learned.

The domain-specific BERT-based models scored higher than the general model on all labels by up to 0.07 points in $F_1$ score. The performance difference was particularly pronounced for the INTV and MEAS labels, which may be due to the former being made up of highly domain-specific tokens, such as generic drug names, and the latter being domain-specific constructions of numbers and units. The label performance differences between the specialised language models were minimal, reflecting those of the overall NER task, with BioBERT having slightly better precision versus SciBERT having better recall.

\begin{table}[h]
    \centering
    \includegraphics[scale=0.5]{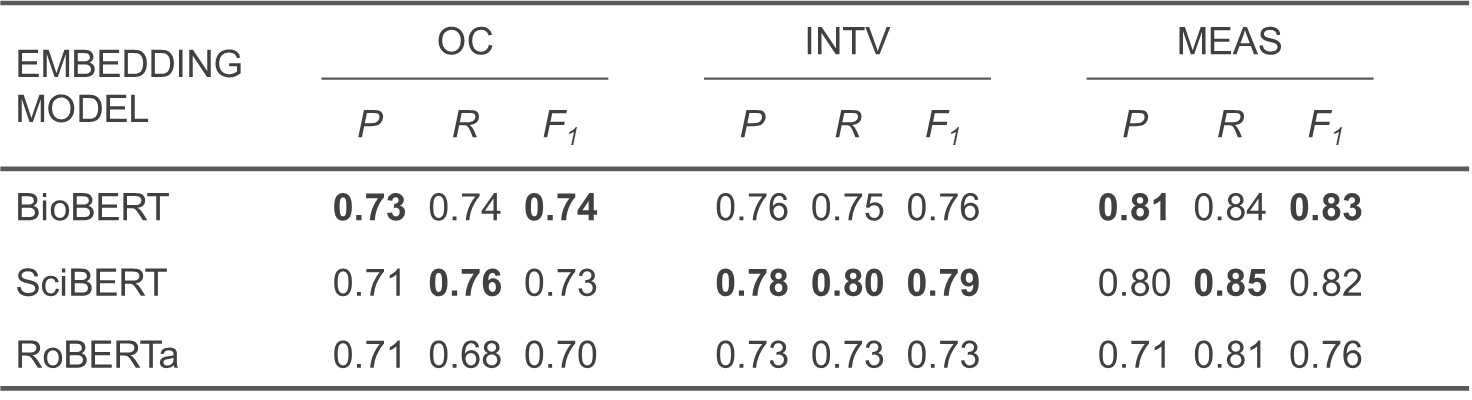}
    \caption{Performance of the NER component for individual entity labels on the all-domains test. Values in bold indicate the language model that scored the highest on a label-specific metric.}
    \label{table:ner_labels}
\end{table}

\subsubsection{RE performance on individual class labels}

We present system performance at classifying the individual RE class labels when given gold standard entities on the all-domains test set in Table \ref{table:rel_labels}. 

Classification performance was highest on the A2\_RES label (linking an arm 2 intervention to its respective outcome measure) at 0.81 $F_1$ score (BioBERT), followed by a similar 0.79 for the A1\_RES label, and a drop to 0.76 for the OC\_RES label (outcome to respective measure). As gold-standard entities are provided to the RE component for evaluating this task, it is unclear whether the performance drop for the OC\_RES label is due to the outcome entity itself or the context linking it to its measure (or, perhaps more likely, a combination of the two). However, the delta between this relation label and the next highest performing is smaller than that of the outcome entity and the next lowest performing NER class label. 

Comparing the language representations, we see similar results to the overall RE performance scores, with RoBERTa performing similarly to BioBERT and SciBERT on all three labels, potentially for reasons outlined previously.  

\begin{table}[h]
    \centering
    \includegraphics[scale=0.5]{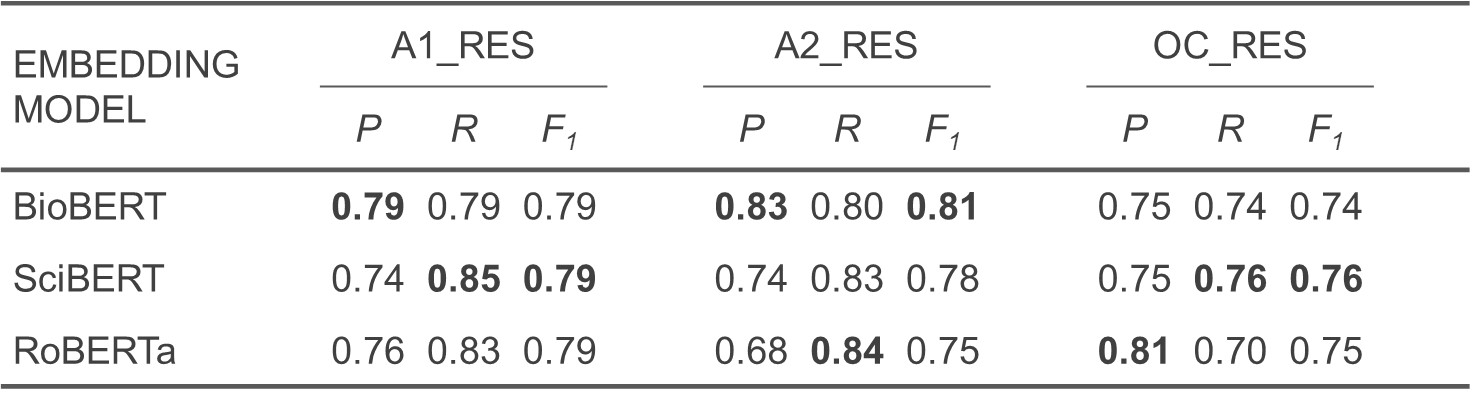}
    \caption{Performance of the RE component for individual relation labels on the all-domains test set. Values in bold indicate the language model that scored the highest on a label-specific metric.}
    \label{table:rel_labels}
\end{table}

\subsection{All-domain performance with respect to the number of training examples}

\begin{figure}[h!]
    \centering
    \includegraphics[scale=0.53]{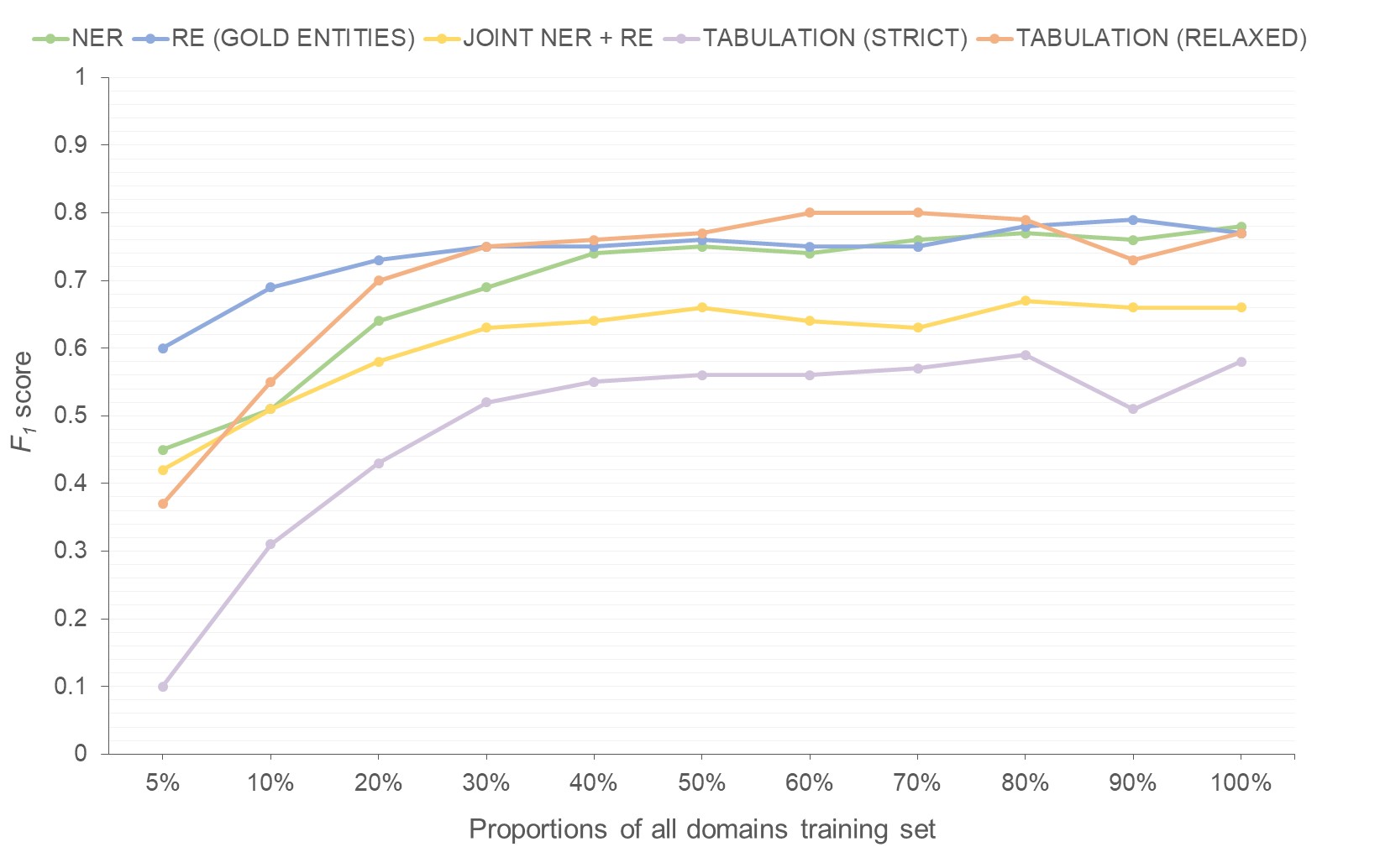}
    \caption{System (BioBERT language model) performance results across the five IE tasks after training on varying proportions of the all-domains training set.}
    \label{fig:strats}
\end{figure}

Figure \ref{fig:strats} shows the impact of varying the number of training examples on the $F_1$ scores of the five evaluation tasks.

This investigation was conducted by taking proportions of the all-domains training set from $5\%$ to $10\%$ and then in increment of $10\%$ to maximum, with an NER and RE component trained on each size stratification. While the SciBERT-based models had slightly higher $F_1$ scores on more tasks, we selected the BioBERT model for this experiment and the out of domain testing in the next section. This was based on the BioBERT model performing better in terms of precision and on the strict tabulation task, with the assumption that false positive tables increase the risk for errors in a systematic review more than empty tables resulting from false negative predictions. 

Across all tasks, $F_1$ score improved as the proportion of the original training set was increased, rapidly from $5\%$ of the set to around $30\%$, where performance gains begins to plateau with diminished returns from $40\%$ to $100\%$. 

The tabulation tasks were the most sensitive to size adjustments in training samples, performing the worst at $5\%$ and having the steepest gradient of ascent as more training examples were introduced. As tasks downstream of both model-based components, this is unsurprising, with errors propagating further errors as they are passed down the pipeline (e.g. one missed entity $\rightarrow$ two related relations missed $\rightarrow$ three related tuples missed). This may explain the drop in performance across all downstream components at $90\%$ training set proportion, with the slight decrease in NER performance at this stratification being amplified in each of the down stream components. Furthermore, as NER $F_1$ score recovers with the full training set, so does that of both the tabulation tasks', even though both RE and joint extraction performance decrease, suggesting NER perturbations have a greater impact on tabulation than those of RE.   

From $10\%$ of the all-domains training set, the relative performance difference between tasks begin to resemble those of the full set. Interestingly, the relaxed tabulation task overtakes both independent tasks (NER and RE on gold entities) from 30 to 80\% proportions, which perhaps results from its differences in performance evaluation in addition to its less restrictive NER matching criteria. 

\subsection{Generalisation performance across domains}

In this section we explore generalisation of our system across disease areas, redistributing the all-domains dataset into train and test sets that are separated by domain (with the development set including the same domains as the train), and retraining our BERT-based (BioBERT) NER and RE components for each experiment. 

\subsubsection{One unseen disease area domain}

\begin{figure}[h]
    \centering
    \includegraphics[scale=0.575]{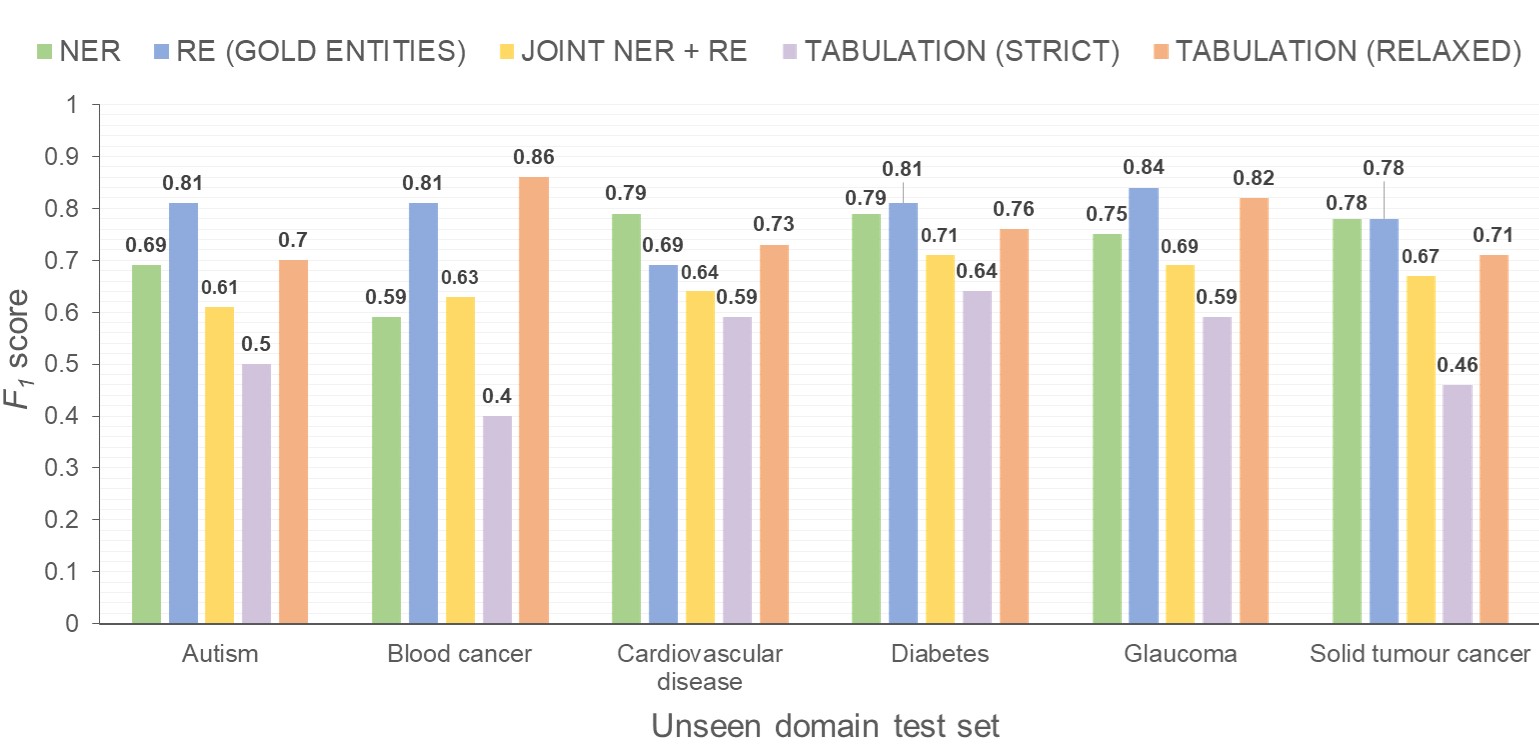}
    \caption{System (BioBERT language model) performance results of the five IE tasks on different unseen disease-area domain test sets.}
    \label{fig:one_unseen}
\end{figure}

Figure \ref{fig:one_unseen} shows $F_1$ score performance on each of the domain areas when removed from the all-domains dataset and used as the unseen test set, with the models being trained on a randomised set of all other domains (e.g. \textit{train set domains:} blood cancer, cardiovascular disease, diabetes, glaucoma; \textit{test set domain}: autism). 
We report relatively comparable performance across domains on all five tasks, with the unseen autism test set having the greatest number of lowest scoring tasks (3). As the least related domain to the other disease areas, this result is unsurprising, especially with the observation that the closely related areas of cardiovascular disease, diabetes and glaucoma are high performers for the system, each domain likely benefiting from the inclusion of the others in the training set. 

Interestingly, the system performs relatively well on its upstream tasks and joint extraction when tested on solid tumour cancer, but struggles with tabulation. The reason for this is unclear, but may be related to how three-way dependencies between measures and their respective outcomes and interventions are constructed in oncology abstracts. 

\subsubsection{Varying the number of training domains}

To explore the importance of training with a domain variety, we trained our system on datasets composed of abstract sentences from a varying number of disease areas.

Three training datasets were composed for this investigation: glaucoma alone (216 randomised examples); glaucoma and cardiovascular disease (108/108 randomised examples); and glaucoma, cardiovascular disease and solid tumour cancer (72/72/72 randomised examples). These domains were selected based on their size, allowing for training sets large enough ($>30\%$ of all-domain set)  to offset the effect of the number of examples on performance.  

\begin{figure}[h]
    \centering
    \includegraphics[scale=0.63]{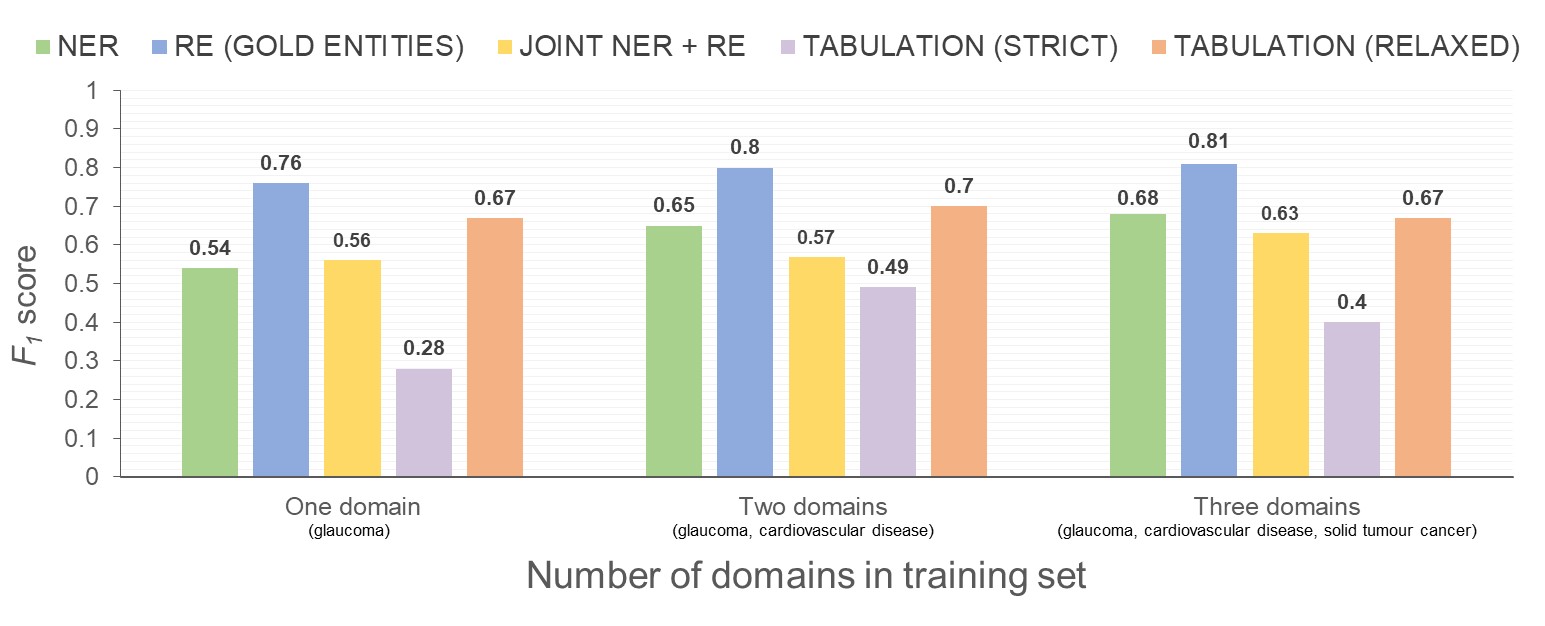}
    \caption{System (BioBERT language model) performance on the autism test set as training domain variety increases.}
    \label{figure:domain_variety}
\end{figure}

We show the results of training on these varying domain sets in Figure \ref{figure:domain_variety}, where the autism test set was chosen as the least-related disease area. From one to two domains, all five system tasks see moderate $F_1$ score increases. However, from two to three domains, only the model-based tasks see a slight performance increase, with both tabulation tasks decreasing in performance. This can be attributed to a shift in recall--precision balance, with recall dropping by almost 0.2 points for the exact tabulation task. This shift was also observed for the joint extraction task, but with precision gains outweighing a relatively tiny change in recall (0.015 point drop), indicating that the tabulation task is highly sensitive to upstream recall ability in this instance. It is unclear why increasing domain variety decreases recall performance -- we leave this to future investigation.

\subsubsection{Comparison of single domain performance}

\begin{figure}[h!]
    \centering
    \includegraphics[scale=0.64]{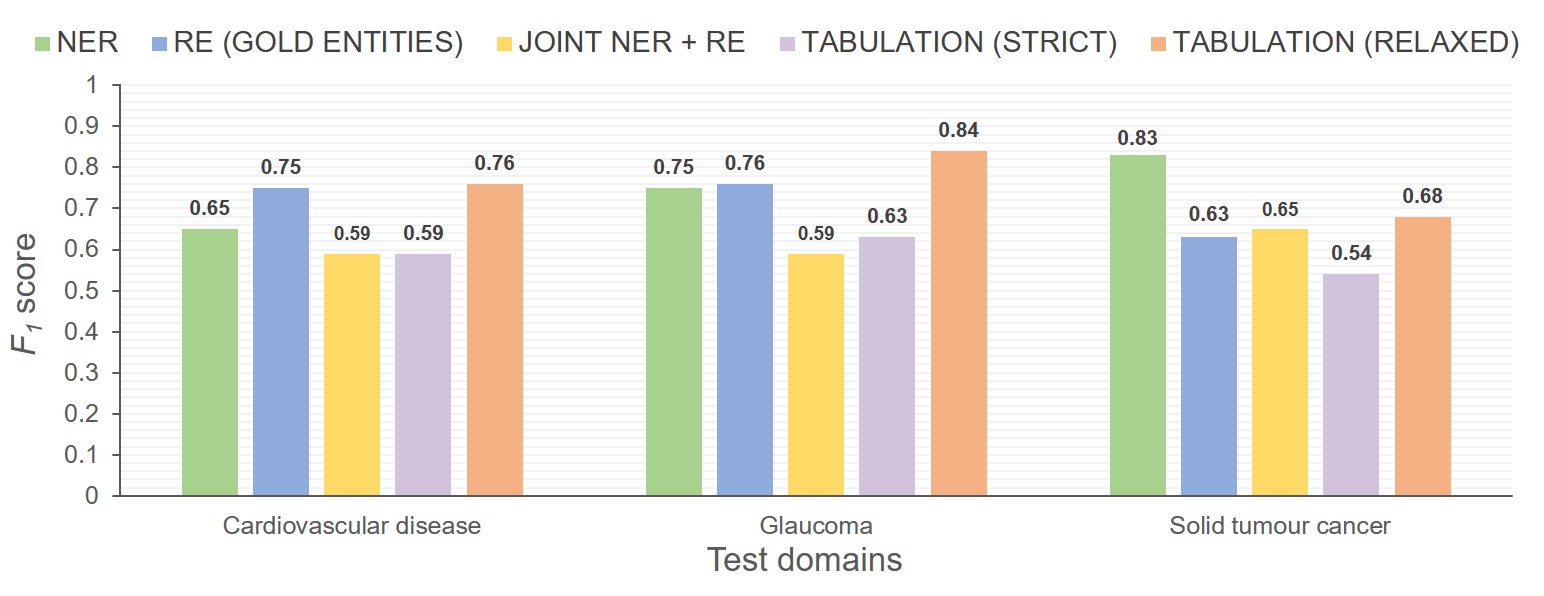}
    \caption{Comparison of system (BioBERT language model) performance after training on different disease areas with the same number of examples, tested on an unseen test set from same domain.}
    \label{figure:single_domains}
\end{figure}

 We compare the performance of our system on individual domain disease areas in Figure \ref{figure:single_domains}.

We again chose the largest three domains for comparison, for the same reasons as outlined in the varying domains section, with each capped to 130 examples (70:10:20 train, dev, test split) for meaningful comparison. 

Glaucoma was the strongest performing dataset out of the three domains, with $F_1$ scores similar to those achieved by the system on the all-domains dataset.

Cardiovascular disease performed somewhat poorly on the NER task (0.65), but achieved relatively good scores across its other tasks. Underlying the low NER performance were difficulties in OC and MEAS label classification, which is in line with the high variability of these classes in this broad disease area, ranging from simple proportional measures of survival to complex patient-reported measures of fitness and exercise. 

Conversely, for solid tumour cancer, the NER component achieved high $F_1$ scores, while those for the RE component were the lowest of the three domains. Again, this makes sense when considering the disease area. In terms of NER, RCT objectives in oncology are often limited to overall survival, progression-free survival, response rate and safety, while many interventions, particularly chemotherapy, have been in use for years and feature repeatedly across studies. However, study structure in oncology is complex, with different combinations of interventions given and tested across cycles, which means relations between objectives, interventions and measures must often be tracked across multiple time points -- a task that is beyond the scope of our system. 

\newpage
\subsection{Error analysis}
\label{sec:error}
In this section, we take a closer look at the common errors occurring within the two main IE models in our system.

\subsubsection{NER errors}

\begin{figure}
\parbox{7.2cm}{
\includegraphics[scale=0.42]{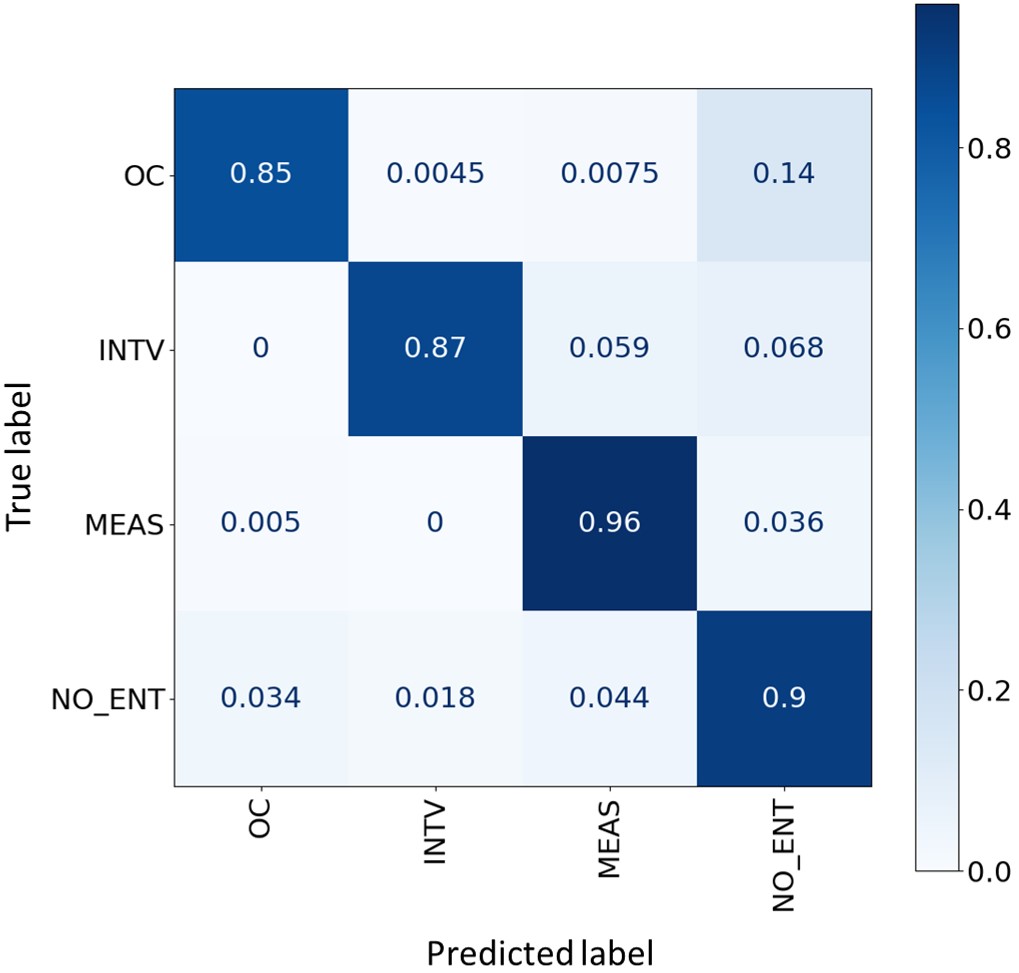}
\caption{Normalised confusion matrix of token-level NER predictions on the all-domains test set.}
\label{figure:ner_cm}}
\qquad
\begin{minipage}{7.2cm}
\includegraphics[scale=0.42]{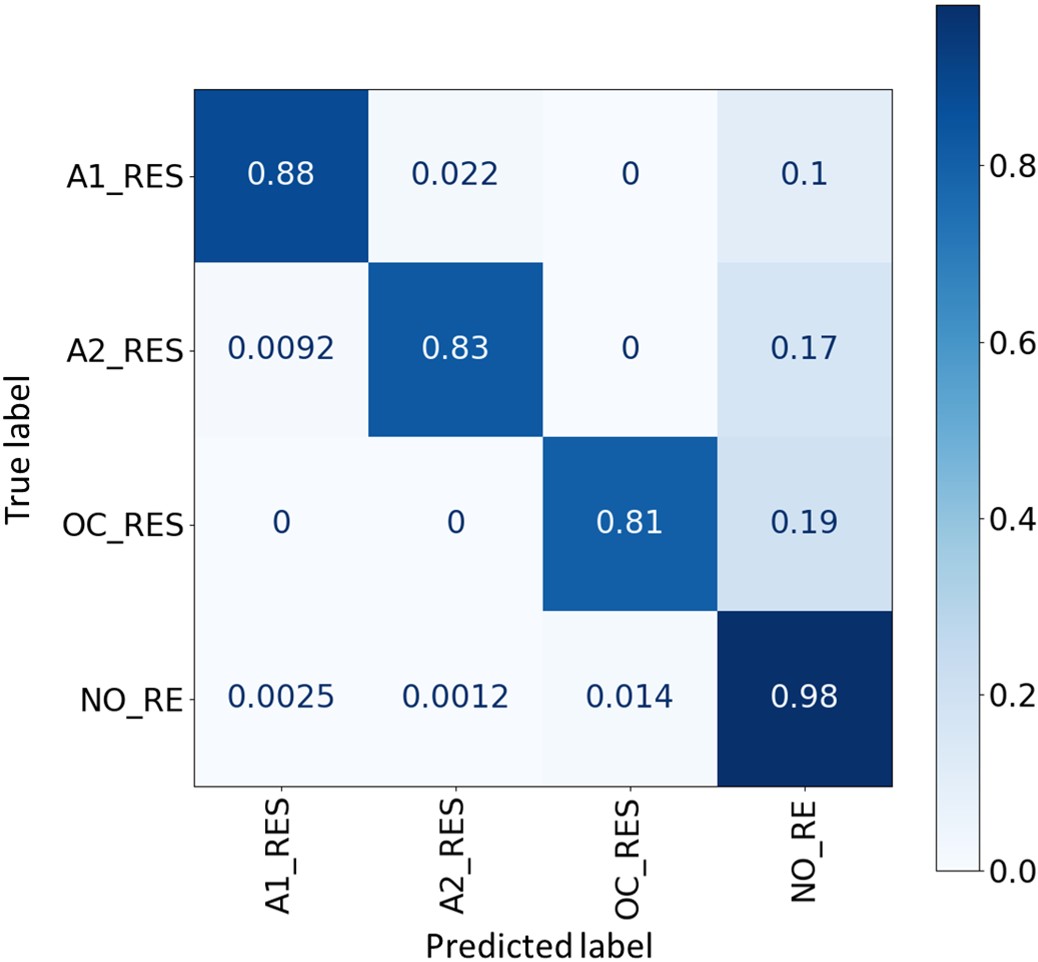}
\caption{Normalised confusion matrix of entity-pair RE predictions on the all-domains test set.}
\label{figure:re_cm}
\end{minipage}
\end{figure}

We present a normalised confusion matrix for our discussion of NER errors in Figure \ref{figure:ner_cm}, which displays token level classifications of the BioBERT-based component on the all-domains dataset. 

\begin{figure}[h]
    \centering
    \includegraphics[scale=0.52]{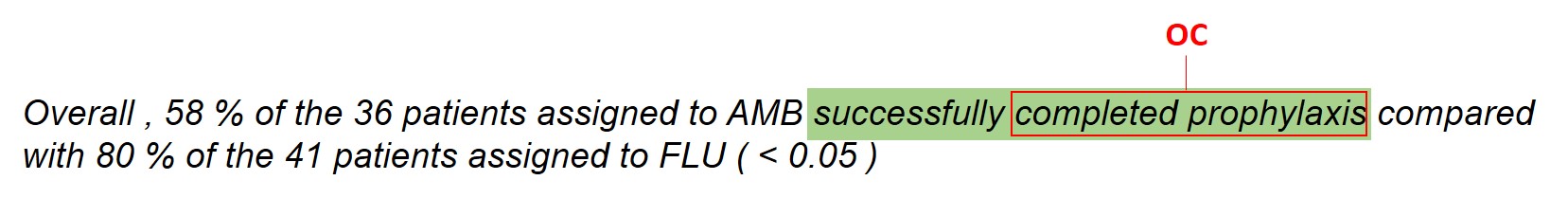}
    \caption{Incomplete OC label error. Gold standard OC labels are in green, while the predicated label is in red.}
    \label{figure:oc_error}
\end{figure}

Incorrect negative classifications are the most common type of misclassification error, particularly affecting the OC label. This is reflected in the commonly observed issue of the model incompletely identifying the tokens in these entities (see Figure \ref{figure:oc_error}), likely due to the reasons discussed in \autoref{subsec:ner_perform}, with outcome entities variable both in length and annotator boundary agreement.

\begin{figure}[h]
    \centering
    \includegraphics[scale=0.52]{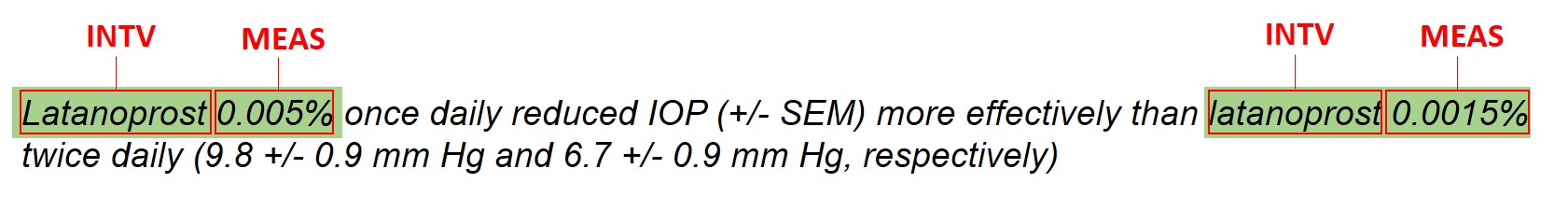}
    \caption{MEAS misclassification error. Gold standard OC labels are in green, while the predicated label is in red.}
    \label{figure:conc}
\end{figure}

Another commonly observed error, reflected in Figure \ref{figure:ner_cm}, is the misclassification of numeric tokens with the MEAS label, primarily occurring with intervention entities which includes concentration values (see Figure \ref{figure:conc}), or non-entity numbers. For the former issue, a post-processing rule that looks one token ahead of interventions for concentration values may be worth investigating. 

\subsubsection{RE errors}

A normalised confusion matrix is also presented for RE errors in Figure \ref{figure:re_cm}, which displays the entity-pair RE classifications of the BioBERT-based component on the all-domains dataset. 

\begin{figure}[h!]
    \centering
    \includegraphics[scale=0.52]{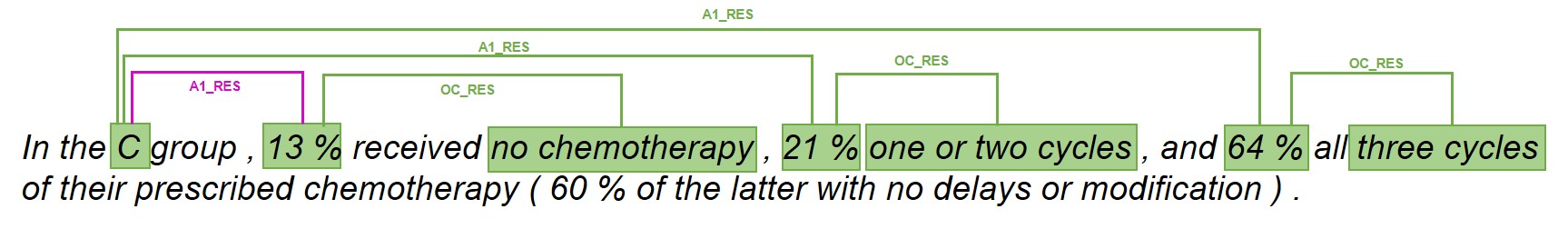}
    \caption{Negative relation classification error. The model only successfully recognises one relation (pink), and misses five (green).}
    \label{figure:rel_error}
\end{figure}

Negative classification of existing relations is the most common type of error, and again occurs most frequently in outcomes and their respective measures. As discussed previously, in disease areas like solid tumour cancer, trial structure can make mapping relations between outcomes and their measures is particularly difficult. In Figure \ref{figure:rel_error} we can see a case of this, where the system fails to categorise a relation between the number of chemotherapy cycles and the proportion of patient who received them.

While reducing the probability threshold of relation classification would reduce the $fn$ error rate, during the implementation stage it was found to impact too greatly on tabulation precision. Presumably, this is because the process of sorting entities through their relations is more sensitive to increases in $fp$ errors, where one incorrect entity results in a $fp$ for the whole tuple.

At the current threshold, positive relationship classes tend not be misclassified as other positive relationships, with no A1\_RES or A2\_RES classes being classified as OC\_RES. Respective arm-1-- and arm-2--measure relationships are occasionally miscast as each other; however, considering the similarity of these relations, this occurs at a lower rate than was expected. 

\section{Conclusion}

In this study, we address a key problem in automating systematic reviews with a system that can tabulate result sentences from published RCT abstracts. Our NLP pipeline achieves this task in three stages: extracting interventions, outcomes and their measures as named entities; identifying the respective relations between them; and using this relational data to sort  entities into the appropriate \textit{outcome}, \textit{arm 1} and \textit{arm 2} columns of an evidence table.

For the two NLP tasks central to our system, NER and RE, we took a transfer learning approach. Through fine-tuning BERT-based transformer models, pre-trained on billions of domain-specific tokens, our system embeds and encodes input sentences into context-rich language representations for these classification tasks. We have also developed an extensive corpus of almost 600 RCT result sentences across six disease areas for training  these models and testing them, as well as our whole system. 

In its primary task of tabulation, our system (BioBERT-based models) achieved an $F_1$ score of 0.58 with strict entity matching and 0.77 with relaxed. If we consider the general NLP tasks of the system against the literature, our $F_1$ scores are relatively high for domain-specific NER (0.78), RE (0.77) and joint NER + RE (0.66), on a test set that includes six different disease area domains; however, the data inclusion criteria of our study must be considered when making such comparisons. In the context of the Trenta et al. \cite{trenta2015extraction} study, from which we derived these criteria and our glaucoma dataset, we see how much the field has progressed thanks to recent contextual language representations. Our pipeline not only differentiates entities that their \textit{one hot vector}-based classification system performed poorly on, such as outcome measures by study arm, but also extracts full entity spans rather than just single-token syntactic heads.

We found that our system generalised well when tested on disease area domains unseen during training, with as few as two domains needed within the training set to achieve this performance. Lastly, our overall results obtained on the all-domains dataset can be achieved by fine-tuning the layers of our models with relatively small training sets of around 200 example sentences. 

\subsection{System limitations}

\textbf{Outcome measure limitations.} Our system has a narrow focus on measure entities, omitting the comparative statistics between arm respective values, which are an essential part of evidence tables. As they have been addressed by other studies, such as that of Kang et al. \cite{kang2019pretraining}, we made a decision at the start of the study to focus only on entities that could be clearly divided into table columns. However, the system could be easily extended to include these measures, with a new comparative statistics entity class that could be related back to the specific outcome entity. 

\textbf{Constraint to the sentence level.} Although interventions, outcomes and their respective measures most frequently occur together in sentences, multi-sentence constructions are not uncommon, and represent a potential blind spot for our system. However, while we chose to operate at the sentence level, this is not an inbuilt limitation of our architecture. BERT language representations have a window of sequence length within which they operate best -- a potential solution to this limitation may be to include as many sentences within this window as possible (without truncating the last inclusion). 

\textbf{Working with long entities.} While by no means an impossible task for our system, longer entities were harder to classify, particularly outcomes containing hierarchies of sub-entities, which were often predicted individually. A solution based on our existing approach may be to decompose these entities into their constitute parts, and share the problem with the RE component. For example, for the outcome entity, \textit{``reduction in interocular pressure of at least 18 mm Hg"} , annotators could label  \textit{``reduction in interocular pressure"} as the main entity, while \textit{``at least 18 mm Hg"} could be tagged as a qualifier, with a relationship mapped between them for the RE component to resolve.

\textbf{Transformers as resource intensive models.} Training and run-time efficiency were not consider during the development of our system. Transformers require high specification GPU hardware to run, and our system includes two of these models. While using a single transformer for both classification models was explored, the observed reduction in performance (overall $f_1$ score drops of up to 0.2) was consider too great for this approach to be of value. However, we would argue the hardware and energy costs related to our system would be outweighed by reductions in the human labour cost of systematic reviews.

\subsection{Future research}

\textbf{Expanding the input scope.} Further investigation into the performance of this system on inputs with a less restrictive inclusion criteria is warranted. In addition to some of the possible explorations outlined in the previous section, future research on this system could investigate extending it to process studies with more than two arms, as well as expanding its extraction scope to all PICO elements from all abstract sentences. 

\textbf{Sentence pairing.} The distinction between between a study's primary and secondary outcomes is an important one for a systematic review, and is usually defined in the methodology section of an abstract. If extended to full abstracts, an interesting component for our system could be a question--answer sentence pairing module (the second task BERT language representations are trained on), linking the primary outcomes defined in the methodology to their respective result sentences. 

\textbf{Going beyond abstracts.} Processing abstracts only represents the first stage of a systematic review. Once a study has been accepted for inclusion, information from the full published paper must be extracted to complete the evidence table. As our system works at a sentence level, it is not inconceivable that it could work well with the sentence from the result section of a full paper, and calls for further investigation. 

\textbf{Argumentation and clinical recommendation.} Considering more advanced areas of research, our system or parts of it, could be used as a component in frameworks for the automation of evidence-based clinical recommendations. More specifically, tabulated result sentences could be used to synthesise claims of the logical arguments that underlie these recommendations.

\subsection{Final remarks}

While a great amount of work still remains in automating systematic reviews of clinical evidence, our study has shown that a key barrier -- differentiating interventions, outcomes and their measures into relevant categories -- may be overcome with context-based language representations, and decomposing the classification problems across a pipeline approach. In the short term, this technology could be used to semi-automate construction of evidence tables, potentially as a first pass process that allows reviewers to start from a pre-filled baseline. Long term, as language representations evolve, and more innovative methods are developed to classify their outputs, it is conceivable that future systems could play an even greater role in automating the systematic review process, with medical domain experts needed only for oversight. This could potentially result in thousands of hours saved in labour costs across the healthcare industry, which could be redirected to achieve the ultimate goal of improving patient care.



\bibliographystyle{plain} 
\bibliography{bibliography}

\end{document}